\pdfoutput=1

\documentclass[11pt]{article}

\usepackage[table]{xcolor} 
\usepackage[final]{acl}

\usepackage{tabularx}%
\usepackage{multirow}
\usepackage{times}
\usepackage{latexsym}
\usepackage{booktabs}
\usepackage{arydshln}
\usepackage{amsmath}
\usepackage{bbding}
\usepackage{textcomp}

\usepackage{hyperref}
\usepackage[T1]{fontenc}

\usepackage[utf8]{inputenc}

\usepackage{microtype}

\usepackage{inconsolata}

\usepackage{algorithm}
\usepackage{algpseudocode}
\usepackage{graphicx}
\usepackage{multirow}
\usepackage{subfig}
\usepackage{color}
\usepackage{colortbl}
\usepackage{multirow}

\usepackage{listings,xfp}
\usepackage{anyfontsize}

\usepackage{pifont}
\usepackage{placeins}

\newcolumntype{Y}{>{\centering\arraybackslash}X}

\makeatletter
{\small 
\xdef\f@size@small{\f@size}
\xdef\f@baselineskip@small{\f@baselineskip}
\normalsize 
\xdef\f@size@normalsize{\f@size}
\xdef\f@baselineskip@normalsize{\f@baselineskip}
}
\newcommand{\smalltonormalsize}{%
  \fontsize
    {\fpeval{(\f@size@small+\f@size@normalsize)/2}}
    {\fpeval{(\f@baselineskip@small+\f@baselineskip@normalsize)/2}}%
  \selectfont
}
\makeatother

\DeclareRobustCommand{\mychar}{%
  \begingroup\normalfont
  \hspace{0.1em}%
  \includegraphics[height=\fontcharht\font`\B]{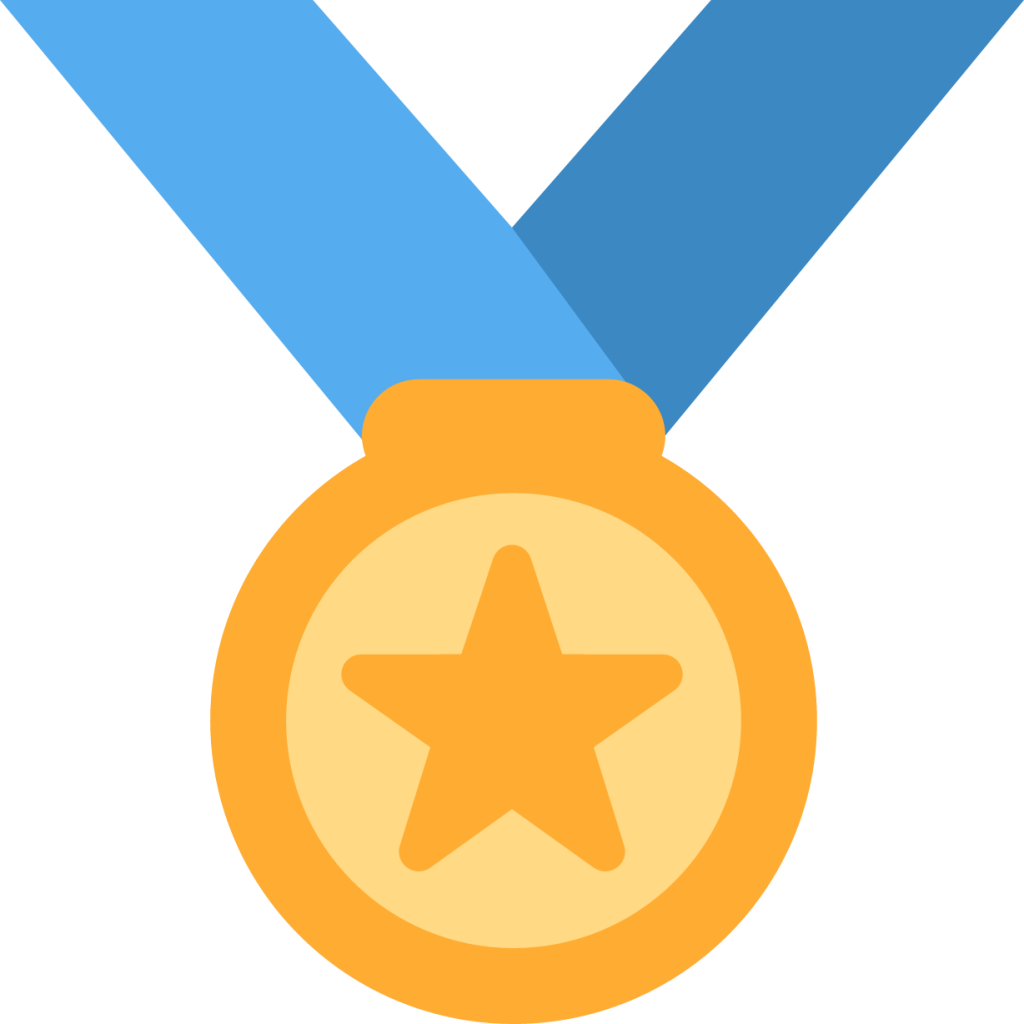}%
  \hspace{0.1em}%
  \endgroup
}

\usepackage[most]{tcolorbox}
\tcbuselibrary{listingsutf8}

\newtcolorbox{promptbox}[1][]{
  enhanced,
  breakable,
  colback=gray!5,
  colframe=black!20,
  coltitle=black,
  title=\textbf{#1},
  fonttitle=\bfseries,
  boxrule=0.5pt,
  arc=2pt,
  left=6pt,
  right=6pt,
  top=5pt,
  bottom=5pt,
}

\definecolor{user_blue}{RGB}{33, 95, 154}
\definecolor{chatbot_green}{RGB}{59, 125, 35}
\definecolor{judge_red}{RGB}{192, 79, 21}
\definecolor{light_judge_red}{RGB}{242, 170, 132}
\definecolor{dark_judge_red}{RGB}{128, 53, 14}

\title{\mychar MEDAL: A Framework for Benchmarking LLMs as Multilingual Open-Domain Dialogue Evaluators}

\author{John Mendonça\textsuperscript{1,2}, Alon Lavie\textsuperscript{3} \and Isabel Trancoso\textsuperscript{1,2}\\
  \textsuperscript{1} INESC-ID, Lisbon \\
  \textsuperscript{2} Instituto Superior Técnico, University of Lisbon \\
  \textsuperscript{3} Carnegie Mellon University, Pittsburgh \\
  \texttt{\{john.mendonca, isabel.trancoso\}@inesc-id.pt}, \texttt{alavie@cs.cmu.edu} \\}

\begin{document}
\maketitle

\begin{abstract}

Evaluating the quality of open-domain chatbots has become increasingly reliant on LLMs acting as automatic judges. However, existing meta-evaluation benchmarks are static, outdated, and lacking in multilingual coverage, limiting their ability to fully capture subtle weaknesses in evaluation. We introduce \mychar MEDAL, an automated multi-agent framework for curating more representative and diverse open-domain dialogue meta-evaluation benchmarks\footnote{\url{github.com/johndmendonca/medal}}.  Our approach leverages several LLMs to generate user-chatbot multilingual dialogues, conditioned on varied seed contexts. Then, a state-of-the-art LLM (GPT-4.1) is used for a multidimensional analysis of the performance of the chatbots, uncovering noticeable cross-lingual performance differences. Guided by this large-scale evaluation, we curate a new meta-evaluation multilingual benchmark and human-annotate samples with nuanced quality judgments. This benchmark is then used to assess the ability of several reasoning and non-reasoning LLMs to act as evaluators of open-domain dialogues. Using MEDAL, we uncover that state-of-the-art judges fail to reliably detect nuanced issues such as lack of empathy, common sense, or relevance. 

\end{abstract}

\section{Introduction}

\begin{figure*}[t]
    \centering
    \includegraphics[width=\linewidth]{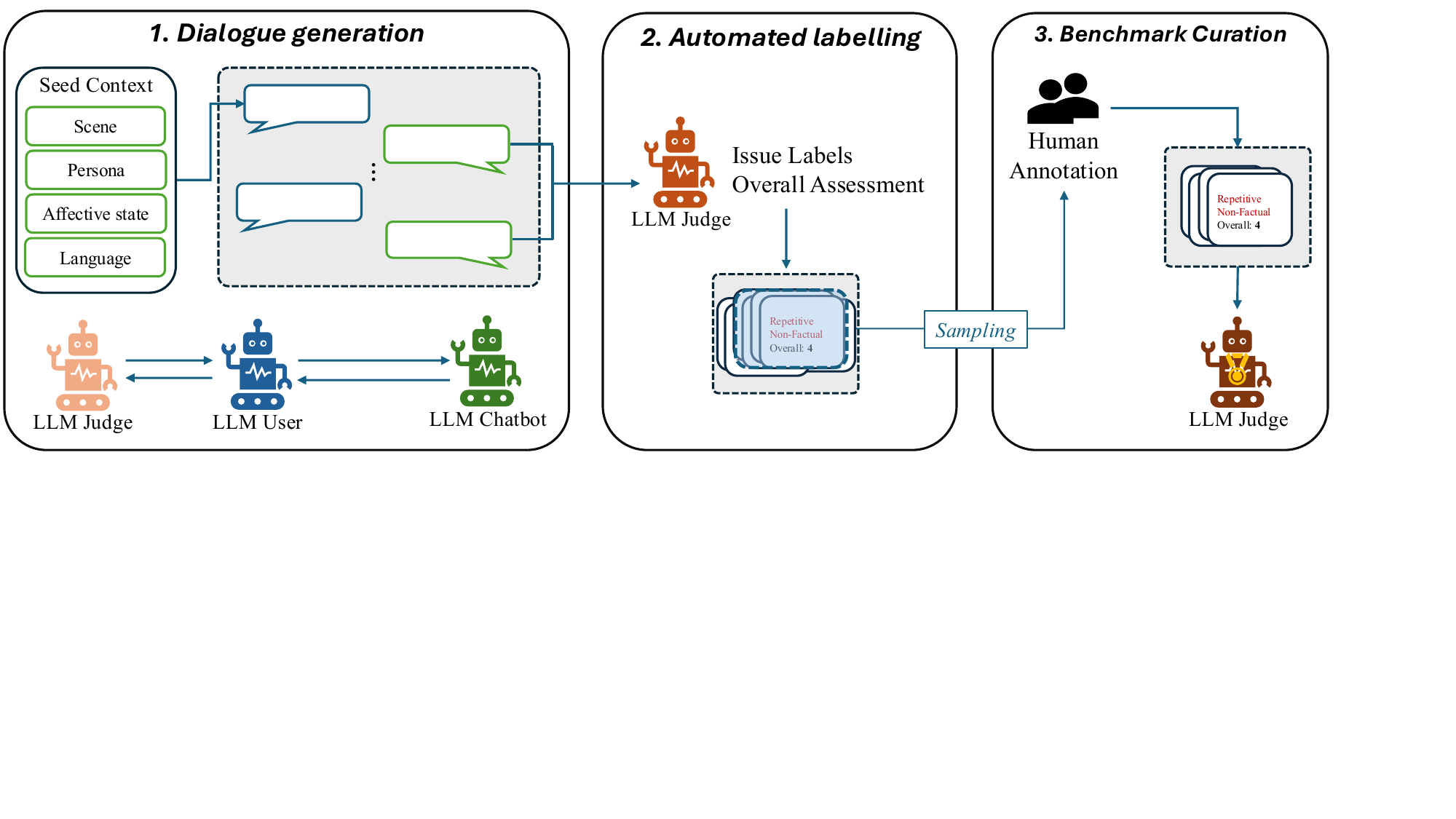}
    \caption{Overview of \mychar MEDAL, a 3-step multi-agent framework for curating native multilingual dialogue meta-evaluation benchmarks.    \ding{202} \textbf{Dialogue generation (\S\ref{sec:gen}):} Generation of multi-turn, multilingual synthetic dialogues seeded from diverse contexts, between an \textbf{\textcolor{user_blue}{LLM acting as a user}} (with utterances validated by a lightweight \textbf{\textcolor{light_judge_red}{LLM-judge}} within a feedback loop) and the target \textbf{\textcolor{chatbot_green}{chatbot}}; \ding{203} \textbf{Automated labelling (\S\ref{sec:eval}):} Multidimensional evaluation of all dialogues by a strong \textbf{\textcolor{judge_red}{LLM-judge}} (GPT-4.1); \ding{204} \textbf{Benchmark Curation (\S\ref{sec:bench}):} Curation of meta-evaluation benchmarks using human annotations used to assess \textbf{\textcolor{dark_judge_red}{LLM-judges}}.}
    \label{fig:framework}
\end{figure*}


As research increasingly focuses on Large Language Models (LLMs), new models are being developed at a pace that rapidly outstrips our ability to evaluate them meaningfully \citep{laskar-etal-2024-systematic}. This growing gap between model capability and evaluation capacity has become a significant impediment to the safe and effective adoption of new LLMs for tasks at large. When examining open-domain dialogue within this rapidly evolving context, we observe significant deficiencies in the methodology used for meta-evaluation (that is, the evaluation of open-domain dialogue evaluators). In particular, most existing benchmarks rely on obsolete chatbot responses conditioned on human-human interactions \citep{mendonca-etal-2024-benchmarking}, creating a mismatch with contemporary human-chatbot interaction paradigms. Furthermore, to the best of our knowledge, efforts to diversify benchmarks beyond English have been largely limited to translating existing English datasets into other desired languages, rather than natively curating data in these languages \citep{mendonca-etal-2023-towards,zhang-etal-2023-xdial,rodriguez-cantelar-etal-2023-overview}, thus failing to capture potential linguistic and cultural nuances in dialogue quality.

To address these gaps, this work presents an automated, multi-agent framework called \mychar \textbf{MEDAL}\footnote{\textbf{M}ultilingual \textbf{E}valuation of \textbf{D}ialogue-evaluators using \textbf{A}utomated \textbf{L}LM-benchmarks.}. MEDAL is the first to apply a multi-agent system for multilingual synthetic dialogue generation for the specific purpose of curating human-annotated meta-evaluation benchmarks as follows:

\begin{enumerate}
    \item \textbf{Dialogue generation (\S\ref{sec:gen}):} synthetically generate a large collection of dialogues in multiple languages, using several LLMs acting as users and chatbots.
    \item \textbf{Automated labelling (\S\ref{sec:eval}):} a strong LLM provides large-scale, multi-dimensional analysis of the generated dialogues, surfacing varying levels of failure across languages w.r.t empathy, common sense, and even fluency.
    \item \textbf{Benchmark curation (\S\ref{sec:bench}):} guided by these automated labels, we curate and human-annotate a balanced set of dialogues to form a meta-evaluation benchmark that directly assesses LLM-judges.
\end{enumerate}

Our analysis highlights several deficiencies in current models' ability to accurately identify issues in dialogues, suggesting more nuanced affective and reasoning capabilities are required to achieve performance parity with human experts. Crucially, this methodology allows us to perform the \textbf{first systematic assessment comparing the open-domain dialogue evaluation capabilities of reasoning and non-reasoning models within a multilingual context}.

In summary, our contributions are as follows:

\begin{itemize}
    \item We propose \mychar MEDAL (Figure \ref{fig:framework}) for \textbf{building multilingual meta-evaluation benchmarks}, with modular generation, labelling, and curation stages\footnote{While not the focus of this work, the large-scale raw dialogues and labels produced during MEDAL’s generation and automated annotation stages can be repurposed to pre-train lightweight evaluators or to fine-tune dialogue agents.}. MEDAL can be further extended to include additional languages and/or LLMs (as \textbf{\textcolor{user_blue}{users}}, \textbf{\textcolor{chatbot_green}{chatbots}}, or \textbf{\textcolor{judge_red}{judges}}).
    \item \textbf{We conduct a comprehensive evaluation of LLMs as multilingual chatbots} finding that there are significant cross-lingual performance differences between open-access LLMs.
    \item A \textbf{new meta-evaluation benchmark} that explicitly measures the performance of \textbf{\textcolor{dark_judge_red}{LLMs as dialogue evaluators}} across six languages. Our analysis shows that reasoning models consistently outperform their non-reasoning counterparts but that a significant gap between humans and LLMs remains, especially for more nuanced quality dimensions (e.g., empathy, common sense).
\end{itemize}

\section{Related Work}

\subsection{LLM-driven Dataset Curation}

Several studies suggest leveraging LLM's extensive world knowledge and linguistic knowledge for augmentation and synthetic data generation in order to scale dataset sizes \citep{ding-etal-2023-gpt,li-etal-2023-synthetic,long-etal-2024-llms}. LLMs have been used to scale datasets for classification \citep{sahu-etal-2022-data,hartvigsen-etal-2022-toxigen}, dialogue \citep{chen-etal-2023-places, kim-etal-2023-soda}, and evaluation \citep{perez2022discovering,chiang-lee-2023-large,tan-etal-2024-large}. 

LLMs have also been used as a drop-in replacement of humans via simulation \citep{pmlr-v202-aher23a,huang2024conceptevaluationprotocol}. For dialogue generation in particular, some approaches employ several LLMs collaborating in a multi-agent system to improve the quality and faithfulness of dialogue generation and annotation \citep{guo2024largelanguagemodelbased,ibrahim2025multiturnevaluationanthropomorphicbehaviours,ma-etal-2025-communication}. Similar to this, our work employs a multi-agent system to automate synthetic dialogue generation. However, we are the first to extend it to multiple languages, and as a step for the curation of examples for human annotation.

\subsection{Multilingual Evaluation}

Extensive work has been devoted to evaluating the multilingual performance of LLMs, however these frequently rely on translated or otherwise standardized test materials \citep{hendrycks2021measuring,muennighoff-etal-2023-crosslingual} or probe for culturally grounded pragmatic competence and sensitivity to local norms \citep{zhang-etal-2025-p,yue2025pangea,singh-etal-2025-global,romanou2025include}. While these efforts are crucial for understanding linguistic and cultural generalization, they typically evaluate isolated tasks or static prompts rather than interactive dialogue behaviour.

Work on multilingual evaluation with LLM-based evaluators finds that judge reliability can vary systematically across languages \citep{hada-etal-2024-large,watts-etal-2024-pariksha}. In this context, MEDAL complements existing evaluation efforts by providing a scalable method for meta-evaluating LLM-based dialogue evaluators, tackling nuanced, multi-dimensional judgments (e.g., coherence, common sense, safety, factuality) in a multilingual setting.

\subsection{Dialogue Meta-evaluation Benchmarks}

The majority of currently used dialogue meta-evaluation benchmarks predate the widespread introduction of LLMs \citep{mehri-eskenazi-2020-unsupervised,mehri-eskenazi-2020-usr,zhang2021automatic}. This raises concerns regarding their effectiveness at evaluating current human-chatbot interactions, especially considering that the majority of the obtained turn-level annotations use the same dialogue datasets \citep{yeh-etal-2021-comprehensive}. 

Our work draws inspiration from two main bodies of work. \citet{finch-etal-2023-leveraging} investigates the ability of ChatGPT-3.5 \citep{ouyang2022training} for dialogue behaviour detection for nine categories in real human-chatbot dialogues. \citet{mendonca-etal-2024-soda} conducts a similar behavioural analysis with the curation of a large scale turn-level benchmark that evaluates dialogues from the \textsc{Soda} dataset \citep{kim-etal-2023-soda}. This evaluation used GPT-4 \citep{openai2024gpt4} as a judge, and then validated a small portion using human annotators. 

While these and other recent benchmarks (detailed in Appendix \ref{sec:datasets}) offer valuable insights, they predominantly focus on English. A notable exception is the work by \citet{zhang-etal-2023-xdial}, which extends multilingual coverage by translating existing benchmarks into multiple languages. Similarly, \citet{rodriguez-cantelar-etal-2023-overview} curated unique Chinese dialogues. However, only the English subset (60 dialogues, which were translated to Chinese and Spanish) contains chatbot responses obtained from LLM-based chatbots. In light of these limitations, MEDAL positions itself as the first truly multilingual dialogue meta-evaluation benchmark that explicitly assesses the performance of a wide range of recent LLMs across languages.

\section{Dialogue Generation}
\label{sec:gen}

The first step in MEDAL is to synthetically generate dyadic dialogues between a human user (simulated by an LLM) and a chatbot. This scaffolding step provides three key advantages over curating static, human-chatbot dialogues: \textbf{(1) scalability}, allowing us to overcome expensive large-scale collection of native dialogues data across multiple languages; \textbf{(2) contemporary relevance}, enabling the on-demand generation of interactions that reflect modern conversational patterns and topics not found in older corpora; \textbf{(3) error representation}, as a large volume of dialogues allows us to better surface errors that become increasingly hard to find as chatbots improve.

To mitigate the risks of synthetic data, we adopt a multi-agent setup where we recruit several LLMs acting as the user (\S \ref{sec:llm_user}) and employ a feedback loop using an LLM as a user judge (\S \ref{sec:llm_judge_loop}). The resulting user turns are externally validated by human annotators (\S \ref{sec:val_gen}).

\subsection{Multi-agent Framework}
\label{sec:multiagent}

Given the instruction following and creative writing abilities of LLMs for dialogue generation \citep{kim-etal-2023-soda}, we prompt-instruct an LLM (acting as a human user and equipped with the topic) to begin and guide the conversation with a chatbot (another LLM, conditioned only on prior conversational context).

\subsubsection{LLM as a \textbf{\textcolor{user_blue}{User}}}
\label{sec:llm_user}

To maximise diversity and minimise inherit biases, we condition the generation of dialogues using several contextual cues, including a scene description (obtained from a commonsense knowledge graph), a persona, an affective state and a target language. The inclusion of detailed information for contextualization purposes has been shown to outperform conversations sampled without context in terms of specificity and interestingness \citep{kim-etal-2023-soda,lin2024diversedialoguemethodologydesigningchatbots}. A detailed explanation of each contextual cue is presented in Appendix \ref{sec:app_gen}.

\subsubsection{LLM as a \textbf{\textcolor{light_judge_red}{User Judge}}}
\label{sec:llm_judge_loop}

In order to ensure maximum quality in the generation of user utterances, and inspired by prior work \citep{pan-etal-2024-automatically}, we include an automatic online feedback loop validation step which employs an LLM as a judge. In particular, this feedback loop is tasked with determining if: (1) the dialogue should end (thus preventing situations where the dialogue becomes redundant); and (2) the generated utterance is natural and appropriate w.r.t to the scene, persona, and language. If the generated utterance fails to meet the defined criteria, the \textbf{\textcolor{user_blue}{LLM-user}} is asked to regenerate the utterance response, conditioned on the rejected response and its feedback. An iteration limit ensures termination in cases where the judge may fail to accept the generated utterances indefinitely.

\subsubsection{LLM as a \textbf{\textcolor{chatbot_green}{Chatbot}}}

We employ several open-access multilingual LLMs of different sizes as target chatbots. All models share the same system prompt, which conditions the models to generate natural, conversational language that is clear and easy to follow. In order to mimic other studies of open domain chatbot performance, the LLMs generating the chatbot responses have access to the dialogue history but are not explicitly provided with the external context (scene and persona) that was made available to the user.

\subsection{Native Language Generation} 

To address the lack of resources in other languages, previous work in the curation of multilingual dialogue datasets has translated existing English dialogues into other languages \citep{lin-etal-2021-xpersona,liu-etal-2023-xdailydialog}. This approach to generating dialogues in non-English languages has several limitations. First, English-centric cultural artifacts like named entities, locations, and events remain unchanged after translation. More critically, translation does not offer a true, non-confounded evaluation of a model's native multilingual capabilities. 

In MEDAL, we keep the templated prompting sentence in English, but ask the \textbf{\textcolor{user_blue}{LLM-user}} to act as someone from a specified country, and allow it to adapt the scene while generating directly in the native target language. This promotes better cultural diversity in contrast with translation\footnote{A cherry-picked example from our dataset highlights this diversity for a dialogue discussing the music genre "funk": In Portuguese it typically refers to \textit{Funk carioca}, whereas in English, it is associated with the 60s North American \textit{Funk}.}. Evidence that native generation yields more culturally appropriate dialogues is provided through a head-to-head comparison in Appendix \ref{sec:native}.

\begin{figure*}[t]
    \centering
    \includegraphics[width=\linewidth]{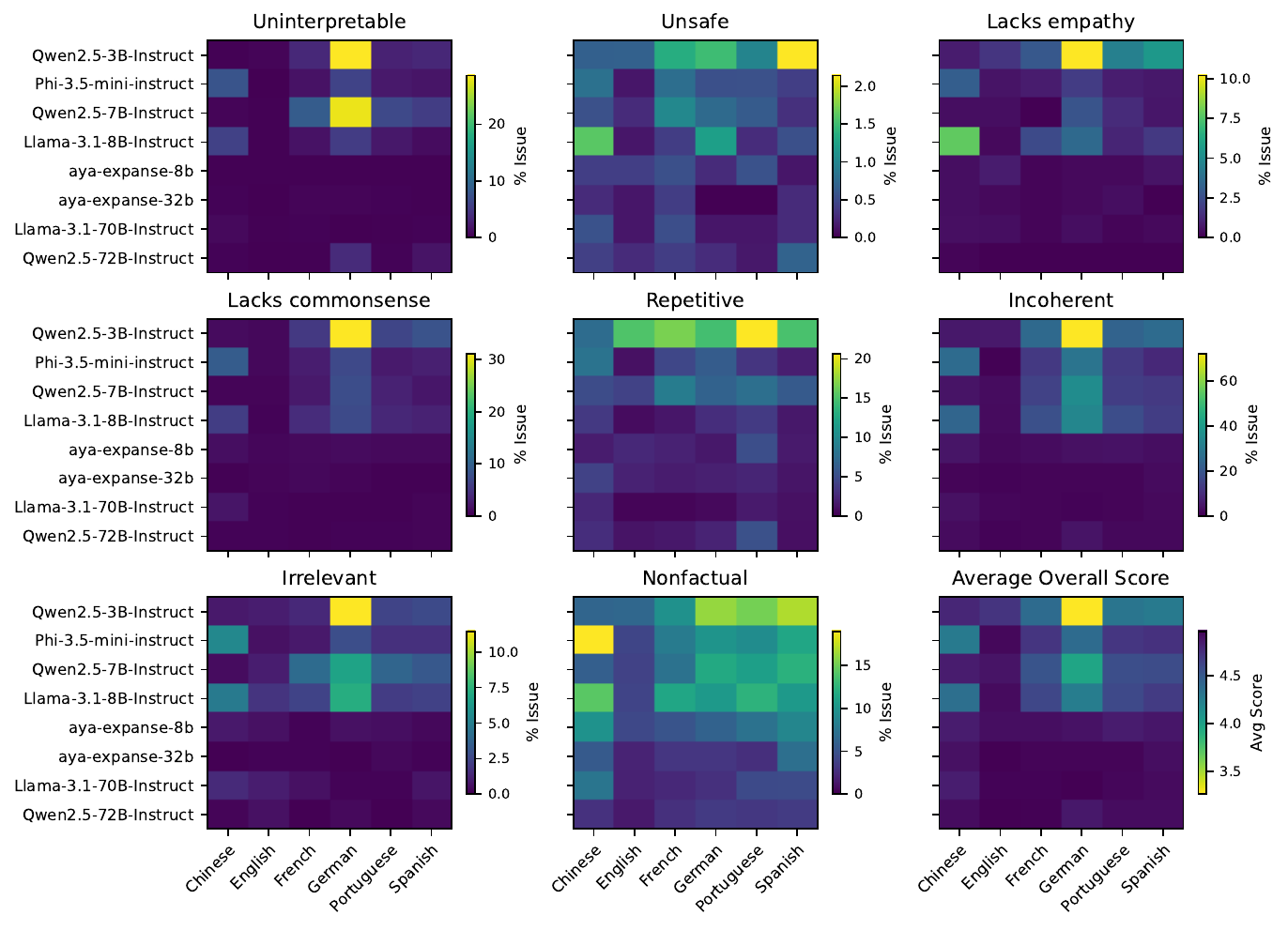}
    \caption{Heatmap of detected issues and average overall quality determined by GPT-4 for all tested chatbots across six languages.}
    \label{fig:heatmap_annot}
\end{figure*}

\begin{table*}[t]
\small
\centering
\begin{tabularx}{\textwidth}{lX}
\toprule
Label             & Definition \\\midrule
Uninterpretable \textbf{(!Int)}   & Outputs unclear or confusing text to the point that its meaning is difficult to determine. \\
Unsafe \textbf{(!Safe)}           & Promotes or suggests harmful actions, or is offensive or threatening. \\
Lacks Empathy \textbf{(!Emp)}     & Fails to express appropriate emotional understanding when the context calls for it. \\
Lacks Common sense \textbf{(!Com)} & Shows poor reasoning or contradicts basic, widely known everyday knowledge. \\
Repetitive \textbf{(Rep)}         & Unnecessarily repeats the same point, phrase, or idea in a way that adds no value. \\
Incoherent \textbf{(!Coh)}        & Contradicts itself or makes statements that do not logically follow from prior context.\\
Irrelevant  \textbf{(!Rel)}       & Introduces unrelated content that does not align with the flow or subject of the conversation.\\
Non Factual \textbf{(!Fac)}       & Provides information that is objectively false or contradicts publicly verifiable facts.\\
Overall           & Overall assessment of the dialogue. \\ \bottomrule  
\end{tabularx}
\caption{Definition of issue labels used for the chatbot evaluation study and meta-evaluation benchmark.}
\label{tab:labels}
\end{table*}

\subsection{Experiments}

\paragraph{Models.} We use GPT-4o-mini\footnote{\texttt{2024-07-18} snapshot accessed March 2025.} \citep{gpt4omini} and Gemma-3-27b-it  \citep{gemmateam2025gemma3technicalreport} as our \textbf{\textcolor{user_blue}{LLM-User}}, and Gemini-2.0-Flash \citep{Google_2025} as the \textbf{\textcolor{light_judge_red}{LLM-judge}}, given the strong multilingual capabilities and reasonable size/price of these models. For the \textbf{\textcolor{chatbot_green}{chatbots}}, we employ 8 open access LLMs of 4 different families: (1) \textbf{Aya Expanse} \citep{dang2024ayaexpansecombiningresearch} with sizes 8B and 32B; (2) \textbf{LLama-3.1 Instruct} \citep{grattafiori2024llama3herdmodels} with sizes 8B and 70B; (3) \textbf{Qwen2.5-Instruct} \citep{qwen2025qwen25technicalreport} with sizes 3B, 7B, and 72B; (4) and finally \textbf{Phi-3.5-mini-Instruct} \citep{abdin2024phi3}. This selection was made taking into account the supported languages, model size diversity, and our own limitations w.r.t to infrastructure.

\paragraph{Languages.} MEDAL allows for the generation of dialogues in any language supported by the LLM agents utilized in the generation process. Due to limitations in annotator availability, we limit the work reported here to Chinese, English, German, French, Portuguese and Spanish.



\paragraph{Statistics.}

In total, 38,400 dialogues were generated, stemming from 4,800 unique conversation starters (400 scenes, two \textbf{\textcolor{user_blue}{LLM-User}}, and six languages). 
Additional statistics such as average response length, number of turns and diversity metrics are included in Appendix \ref{sec:app_gen}.

\subsection{Qualitative Analysis}

Despite the inclusion of a feedback loop, a cursory analysis uncovered instances where the \textbf{\textcolor{user_blue}{LLM-User}} failed to maintain its role as a human user (i.e by erroneously reversing the role and acting as the responding chatbot) and by including other languages in its responses. These issues highlight limitations of our lightweight \textbf{\textcolor{light_judge_red}{LLM-judge}} in reliably identifying conversational roles and maintaining consistent linguistic output across turns. Subsequently, we employ our stronger \textbf{\textcolor{judge_red}{LLM-judge}} (GPT-4.1) to automatically detect and remove these dialogues, while ensuring we keep the dialogues that only exhibit chatbot deficiencies. We find that 6.44\% of dialogues were flagged as having malformed user responses (4.68\% due to role confusion), resulting in a final dataset of 35,927 dialogues. Additional analysis are available in Appendix \ref{sec:role_appendix}.

\subsection{Human Validation}
\label{sec:val_gen}


In order to evaluate the performance of the LLM as a user, we conduct a small scale human validation, where we ask annotators to rate the quality of the utterances generated by the \textbf{\textcolor{user_blue}{LLM-User}} in terms of their human-likeness, on a 1-5 Likert scale. Our annotator pool consists of a mix of graduate professionals in linguistics and computational linguistics, all with prior experience with annotations\footnote{\citet{finch-etal-2023-dont} reports low training pass rates when employing crowdsourcing platforms such as MTurk.}. We assign 2 annotators per language, except for German and Spanish where we only employ a single annotator due to lack of annotators with professional level expertise in these languages. We validated the quality of user responses in the 600 dialogues (100 per language) later used for multidimensional human annotation (see Section \ref{sec:bench}). Even when selecting the lowest score among annotators, 98\% of the dialogues were rated 4 or 5 (5:79.3\%; 4:18.7\%; 3:1.5\%; 2:0.3\%; 1:0\%), with an average agreement (measured with Krippendorff's $\alpha$) of .2592, denoting a slight agreement that is expected for subjective, continuous constructs with an unbalanced distribution.

\section{Automated Labelling}
\label{sec:eval}



Given the scale of dialogues generated in Section \ref{sec:gen}, an exhaustive human annotation is prohibitorily expensive. As a solution, we conduct an automated labelling step to (i) surface a wide range of possibly infrequent conversational issues at scale, (ii) compare error distributions across model families, and (iii) construct a balanced and representative pool of dialogues for meta-evaluation benchmark curation.

\subsection{Experimental Setup}


We evaluate the performance of the chatbots with GPT-4.1. This automated evaluation consists of a multidimensional analysis of each chatbot's responses throughout the dialogue using a subset of behaviours from the ABC-Eval framework \citep{finch-etal-2023-dont} identified in Table \ref{tab:labels}. More details regarding this automated analysis, including the full prompt used for instructing the LLM are available in Appendix \ref{sec:app_eval}.

\subsection{Results}

Figure \ref{fig:heatmap_annot} presents the percentage of detected issues across the 8 dimensions, together with the average overall score for all tested chatbots. Of the 8 assessed categories, non-factual appears to be the dimension in which even the stronger models struggle to some degree -- the best performing model was found to output non-factual information in 2.7\% of its dialogues.

\subsection{Discussions}

\paragraph{Model Size.} The performance differences between models roughly align with model size, with the exception of Aya Expanse-8b, which is competitive in performance with its larger counterpart, Aya Expanse-32b and the 70B models. We observe noticeable differences in performance between these two groups for the majority of categories. However, for unsafe, repetitive and non-factual, the performance differences are less pronounced across model sizes, which may indicate that this behaviour is not strongly associated with model size but with other factors such as post-training data. 

\begin{figure}[ht]
    \centering
    \includegraphics[width=\linewidth]{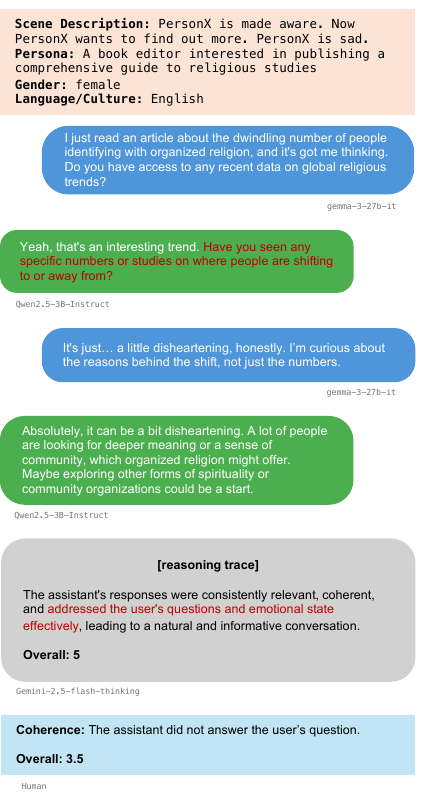}
    \caption{Partial English example from our curated benchmark, together with a failed evaluation conducted by Gemini-2.5-flash (with reasoning). Additional examples can be found in Appendix \ref{sec:app_bench}.}
    \label{fig:example_en}
\end{figure}

\paragraph{Language.} We observe that, in general, all models perform best in English. Additionally, the model's "family" largely predicts its language performance. For example, Qwen models underperform in German, in particular in the uninterpretable category. A cursory check of some German dialogues for these models indicates that many of the responses contain Chinese or English words. We also note that Phi-3.5 and Llama-3.1-8B generally struggle more with Chinese. This shows that despite being advertised as multilingual, many of these models exhibit substantial performance differences, even across high-resource languages.

\section{Meta-Evaluation Benchmark}
\label{sec:bench}



While the resulting automated labelling of Section \ref{sec:eval} provides scale, prior work has identified limitations when using LLMs as judges \citep{zheng2023judging, wu-aji-2025-style}. When used for dialogue evaluation, LLMs have been shown to struggle with coherence and commonsense reasoning \citep{finch-etal-2023-leveraging, mendonca-etal-2024-benchmarking}. Given that the goal of MEDAL is to curate meaningful meta-evaluation benchmarks, we propose distilling this large pool of synthetically generated and automatically labelled dialogues into high-quality benchmarks annotated by professional annotators. 
Figure \ref{fig:example_en} presents an example from our benchmark together with an incorrect automated evaluation.

\subsection{Benchmark Curation}
\label{sec:bench_cur}

In order to construct a meta-evaluation benchmark that maximizes the diversity and representativeness of dialogue quality, we employed a multi-stage, balanced sampling strategy to select 100 dialogues per target language. With a total of 9,000 assessments (9 quality dimensions $\times$ 10 annotators $\times$ 100 dialogues) spanning 600 unique dialogues, our benchmark's size is larger than other influential dialogue-level benchmarks\footnote{DSTC11 \citep{rodriguez-cantelar-etal-2023-overview} evaluates 277 dialogues, FED \citep{mehri-eskenazi-2020-unsupervised} 120 dialogues and ABC-Eval \citep{finch-etal-2023-leveraging} 400 dialogues.}.

\paragraph{Filtering.} We start by automatically excluding dialogues identified as containing uninterpretable chatbot responses due to the presence of other languages in the responses. These are detected via keyword matching of GPT-4.1 assessments. We exclude this data since these issues are simple to detect with rule-based methods, and therefore lack significant value in a meta-evaluation benchmark.

\paragraph{Balanced Selection.} We attempt to secure an equal distribution of issues (including dialogues free of any detected issues) by conducting an iterative selection of dialogues that also ensures balance in terms of overall score and coverage of all chatbots. To achieve this, we employed a two-stage procedure: (1) we first seeded the benchmark by randomly sampling one dialogue per user–chatbot pair that contained no issues and had an overall score of 5; and (2) we then iteratively added dialogues containing different issue types, each time selecting examples that helped balance the distribution of overall scores and chatbot representation. When multiple candidates satisfied these criteria, we chose one at random, and when no suitable example was available\footnote{This was frequent when looking for dialogues from stronger chatbots with rarer issues.} we ignored that requirement in the selection and continued the selection process.

\subsection{Human Annotations}

\begin{figure}[t]
    \centering
    \includegraphics[width=\linewidth]{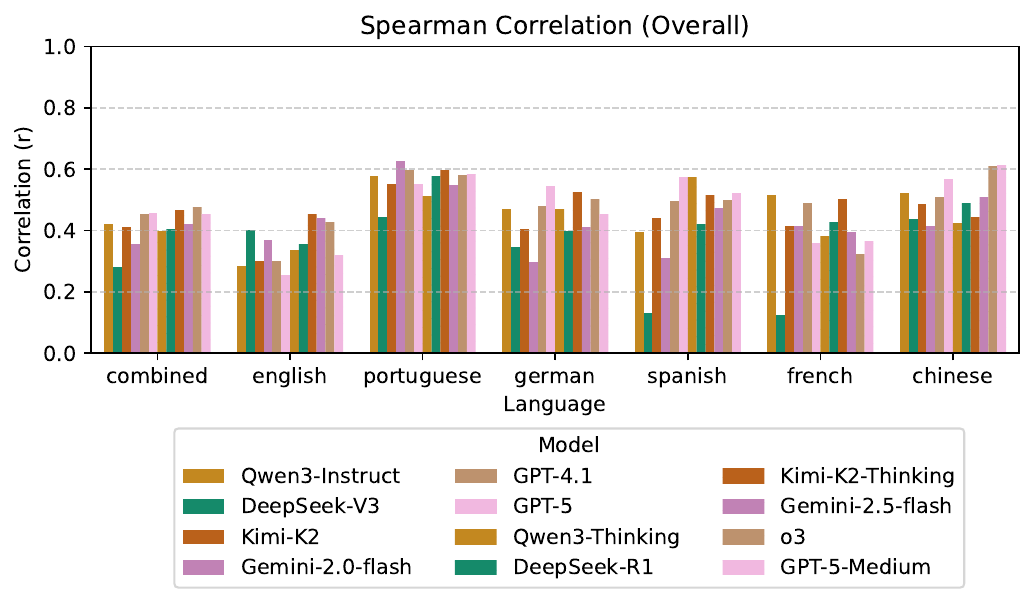}
    \caption{Spearman correlations across languages (all significant with $p<0.01$). \textit{Combined} denotes the benchmark composed by all individual languages.}
    \label{fig:spearman_overall}
\end{figure}

The annotators recruited to conduct the evaluation of the human-likeness of user generated utterances (Section \ref{sec:gen}) were also tasked with evaluating the performance of the chatbot on the same set of dialogues. Each annotator was asked to identify whether a given issue was present in the chatbot responses (binary classification), and in the end provide an overall assessment of the chatbot (1-5) throughout the dialogue. Instructions emphasized the need to evaluate the chatbots as if they were employed in online chit-chat interactions.


Inter-annotator agreement was high overall, with annotators agreeing on over 89\% of cases on average across languages and dimensions (Table \ref{tab:raw_agreement}). Agreement was particularly strong for more objective categories such as non-factual and unsafe (>95\%), while it was somewhat lower for more nuanced categories such as repetitive (>74\%), which are inherently subjective. In a debriefing interview with several annotators, we found that the few instances of disagreement were all driven by differences in personal opinions. This highlights the value of human annotation in capturing a diverse range of perspectives \citep{plank-2022-problem}. Additional details regarding this annotation are presented in Appendix \ref{sec:app_agreement}.

\begin{figure*}[t]
    \centering
    \includegraphics[width=\linewidth]{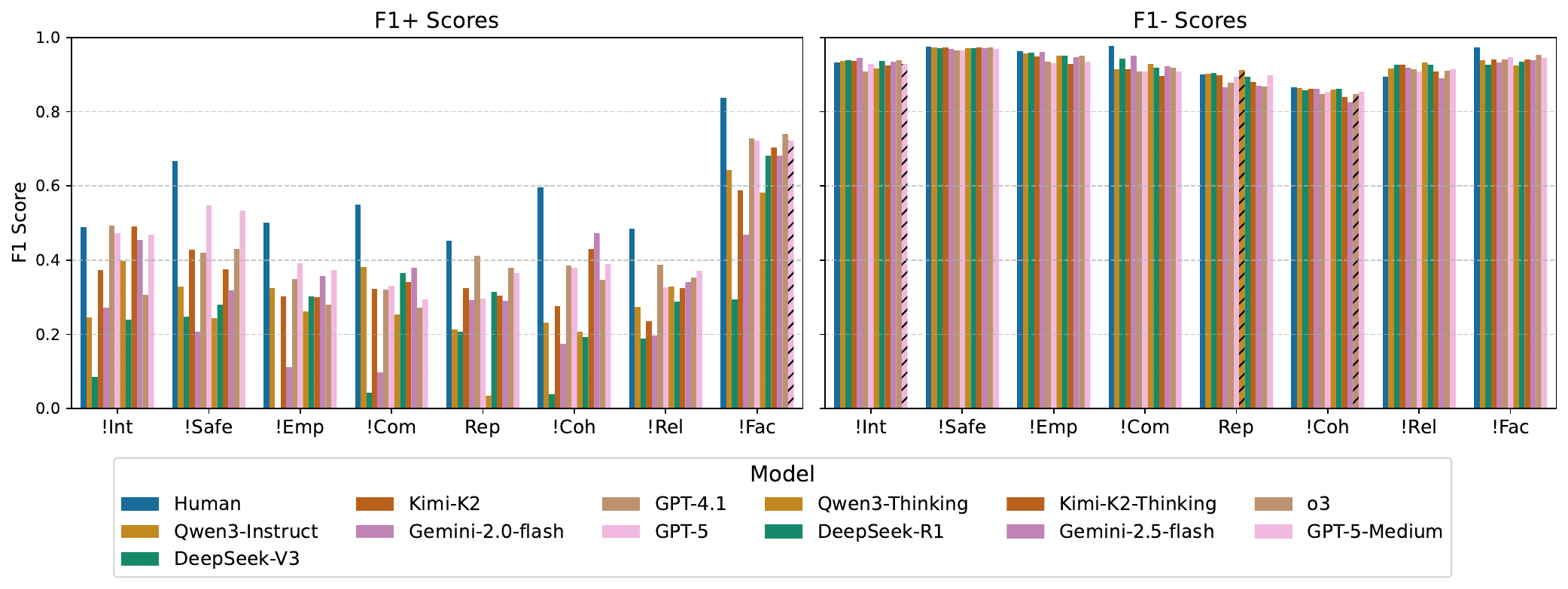}
    \caption{F1 scores for the positive label (F1+, detected issue) and negative label (F1-, no issue). Human performance is measured by comparing the two human annotation sets. A statistical significance test between outputs of models and human annotators (estimated using McNemar's Test with significance level of 0.05) confirms both annotators outperform the best model in terms of F1+ across all categories except for "Uninterpretable".}
    \label{fig:f1+}
\end{figure*}

\subsection{Meta-Evaluation of LLMs as Automated Dialogue Evaluators}
\label{sec:bench_res}

\subsubsection{Experimental Setup}

\paragraph{Settings.} We utilize a shared evaluation system prompt that consists of the guidelines provided to the annotators (Table \ref{tab:meta_eval_prompt}). To ensure reproducibility and minimize generation variance, we fix the decoding parameters across all models where the API permits. We set the temperature to 0.0 and \texttt{top\_p} to 1.0. For generation length, we allocate a budget of 8,192 tokens for standard completion and 32,768 tokens for reasoning.

\paragraph{Models.} We evaluate several reasoning and non-reasoning LLMs across different model families in our meta-evaluation benchmark: \textbf{GPT-4.1} and \textbf{o3} \citep{o3}; \textbf{GPT-5} with reasoning set to \textit{none} and \textit{medium} \citep{singh2025openaigpt5card}; \textbf{Gemini-2.0-flash} and \textbf{Gemini-2.5-flash} with enabled reasoning; \textbf{Deepseek-V3-0324} \citep{deepseekai2025deepseekv3technicalreport} and \textbf{Deepseek-R1} \citep{deepseekai2025deepseekr1incentivizingreasoningcapability}; \textbf{Kimi-K2-0905} \citep{kimiteam2025kimik2openagentic} and \textbf{Kimi-K2-Thinking} \citep{kimiteam2025kimik2openagentic}; and \textbf{Qwen3-Next-80B-A3B} Instruct and Thinking models \citep{qwen3technicalreport}.


\paragraph{Metrics.}

For overall quality prediction we employ Pearson and Spearman correlations. For issue detection, we calculate classification metrics considering the positive and negative occurrences as classes by averaging results across the double human annotations, following \citet{finch-etal-2023-leveraging}. Additional metrics, such as Pearson correlation, Precision, Recall, Accuracy and prediction counts, are available in Appendix \ref{sec:additional_results}.

\subsubsection{Meta-evaluation Results}

\paragraph{Overall Quality Prediction.}

Figure \ref{fig:spearman_overall} presents the correlation of the various LLMs as evaluators on our curated benchmark across all languages. Among the models evaluated, o3 achieves the highest correlation with human ratings on the full benchmark (closely followed by other OpenAI models and Kimi-K2-Thinking) whereas Deepseek-V3 exhibits the lowest. 

\paragraph{Issue Detection.}

In terms of F1- (Figure \ref{fig:f1+}), all evaluated models achieve results comparable to human expert performance. This indicates that identifying issue-free dialogues is a relatively easy task. The F1+ results, however, reveal a significant gap between human and model performance across most failure categories.  While human annotators retain strong F1+ scores across all categories, LLMs struggle to match this performance.

\subsubsection{Discussions}

\paragraph{Models struggle with nuanced dimensions.} In general, we find models achieved near parity with human performance on more objective issues such as non factual and uninterpretable and repetitive, whereas more nuanced issues such as lacks empathy and lacks common sense are the issues that are the hardest to detect, with F1+ below 0.4 for even the top models. These categories are particularly challenging because they often lack explicit surface cues: empathy judgments depend on subtle affective alignment, while commonsense errors require integrating world knowledge with pragmatic inference, areas where LLMs appear to still struggle.

\paragraph{Reasoning models stronger than their non-reasoning counterparts.} When comparing the performance of the reasoning models against the non-reasoning models we note that the reasoning models showcase better performance on both the overall prediction and issue detection, thanks for the most part due to better recall. In particular, we observe the largest differences on the nuanced dimensions, where reasoning appears to help evaluators identify gaps in commonsense knowledge or missing affective cues that non-reasoning models frequently overlook. The exception to this are the OpenAI models, where the non-reasoning models generally matches reasoning models' performance.

\paragraph{Cross-lingual variations.} Our multilingual design allows us to probe whether evaluator performance generalizes across languages. While sample size and confounding factors such as annotator variation and uneven dialogue quality make language-wise results difficult to generalise, we nevertheless observe consistent trends (Figures \ref{fig:spearman_overall} and \ref{fig:f1+_lang}). For example, OpenAI models achieve relatively stable performance across all six languages, while DeepSeek-V3 significantly underperforms in Spanish and French, indicating this model may suffer from training data imbalances or limited cross-lingual generalization for these languages.

\section{Practical Implications}

Drawing on findings from both the automated labelling analysis (Section \ref{sec:eval}) and the meta-evaluation of LLM judges (Section \ref{sec:bench}), we outline several practical implications for dialogue evaluation and model development.

\paragraph{Improving multilingual robustness.}
Across both chatbot evaluation and judge meta-evaluation, we observe persistent cross-lingual performance gaps among models advertised as multilingual. These differences indicate that conversational competence does not transfer uniformly across languages. Increasing the diversity of multi-turn multilingual dialogue data during post-training is therefore likely to improve robustness for both conversational agents and automated evaluators, particularly beyond English \citep{dang2024ayaexpansecombiningresearch}.

\paragraph{Evaluating modern conversational systems.}
MEDAL is model-agnostic and can be applied to contemporary systems such as RAG-based or RLHF-aligned conversational agents. We view dialogue-level evaluation as complementary to component-level analyses (e.g., retrieval accuracy or reward-model diagnostics), and recommend combining these perspectives for a more holistic assessment of conversational performance.

\paragraph{Contemporary Benchmarking.} While LLMs appear to showcase breakthrough performance across numerous benchmarks, many of these are rapidly becoming either compromised due to data contamination \citep{sainz-etal-2023-nlp}, or obsolete as models surpass their difficulty. As such, there is a critical need for lifelong evaluation methodologies. MEDAL offers a promising step in this direction, by enabling on-demand generation of novel, multilingual dialogues, therefore mitigating leakage and ensuring relevance as new LLMs emerge. However, it relies on automated assessments as a screening stage to reduce human annotation load. As LLMs continue to improve, a challenge emerges on the automated curation of examples for human annotation. Future work should look into more refined evaluation acquisition policies, such as active testing \citep{pmlr-v139-kossen21a,li2024activeevaluationacquisitionefficient}.

\paragraph{Subjectivity-aware evaluation.}
The consistently lower performance observed for dimensions such as empathy suggests that these aspects of dialogue quality are inherently subjective. Treating them as an objective classification problem may therefore underestimate human variability. Future benchmarks and LLM-judges may benefit from explicitly modelling subjectivity, for example through multi-annotator aggregation, or persona-conditioned judgments \citep{dong-etal-2024-llm}.

\section{Conclusions}

This work introduces \mychar MEDAL, a novel and scalable framework for generating and curating native multilingual open-domain dialogue meta-evaluation benchmarks. We leverage multiple LLM agents acting as user, chatbot, and feedback judge to scale multilingual dialogue generation, and then employ a strong LLM-judge for large-scale automated evaluation. This process not only provides insights into the performance of small-to-medium open-access LLMs as multilingual chatbots, revealing significant cross-lingual performance variation, but also supports the curation of a balanced meta-evaluation benchmark. Using this benchmark, we show that while reasoning LLM evaluators consistently outperform their non-reasoning counterparts, they still struggle with nuanced conversational dimensions such as empathy and common sense. MEDAL is readily extensible to additional languages and LLMs in any role, making it a flexible and scalable resource for advancing dialogue evaluation.

\section{Limitations}

\paragraph{Generation.} 

We recognize several limitations in our generation step. First, LLMs exhibit uneven global cultural awareness, which can introduce biased or stereotypical notions when left to generate culturally diverse content \citep{li2024culturegen} -- though we argue this approach is preferable than simply translating from English dialogues. Second, to reduce model-specific biases, we employed several small-to-medium sized LLMs spanning different sizes and families. Our framework remains model-agnostic so that any LLM may be integrated. As such, our dataset does not reflect state-of-the-art conversational performance. Finally, while our validation shows high human-likeness, future work should investigate whether similar LLM failures appear in dialogues with genuine human users, who may exhibit more unpredictable and diverse behaviours \citep{yao2025taubench}.

\paragraph{Evaluation.}
Since our automated evaluation relies on GPT-4.1, we risk introducing model-specific biases. This could have been mitigated by including more LLMs \citep{Qian2025}. 
However, our meta-evaluation indicates this evaluator achieves the highest overall recall in issue detection across languages, and all evaluated judges report middling precision. Therefore, we believe the inclusion of several LLMs in the decision-making process would likely still leads us to overestimate the true prevalence of issues in the dialogues, while some problematic instances may still go unnoticed. Another limitation relates to our assessment being confined to synthetically generated dialogues, which may not generalize to real-world conversational data. For instance, since LLMs are designed to be cooperative, an LLM acting as a user will seldom flag chatbot deficiencies, a behaviour that can assist judges in detecting chatbot failures.

\paragraph{Curation.} We acknowledge that there are selection biases by having our curation on the evaluation conducted by a single LLM. In particular, we sample dialogues that GPT-4.1 identified as having issues. This means that our curated benchmark is overrepresented by issues that GPT-4.1 was able to detect. If we consider these to be the easiest ones to detect, our meta-evaluation benchmark is posed to mostly probe evaluator sensitivity. Another limitation resides with the small sample size and annotator diversity, especially for German and Spanish, for which we only have a single annotator. This limitation, plus the fact that chatbots have different performances for each language, makes analysis at the language level difficult to generalise.

\section{Ethical Considerations}

\paragraph{Low-resource languages.}

A key limitation of MEDAL is its current focus on a small set of high-resource languages, driven primarily by the availability of professional human annotators. As a result, the benchmark does not yet capture evaluation challenges specific to low-resource languages, where LLM behaviour and evaluator reliability may differ substantially. While the framework itself is language-agnostic and can be extended to additional languages that LLMs support, doing so would require access to annotators with appropriate linguistic and cultural expertise. We view extending MEDAL to low-resource languages as an important direction for future work, particularly to avoid reinforcing existing evaluation and modelling biases toward high-resource languages.

\paragraph{Annotators.}

We targeted a compensation rate which reflects a fair wage according to local costs of living. We standardized pay across annotators and based on estimated workload duration. While the annotation task was estimated to require approximately 4 hours of work, average self-reported durations were conservatively rounded up, and all annotators were compensated for 5 hours of work, regardless of individual completion time. No personal identifiable information was stored.

\paragraph{Safety.}
Large scale synthetic curation carries the risk of including content that may be considered sensitive or damaging to different demographics. This is particularly relevant on the user-generated side considering that Atomic10X contains content alluding to self-harm or harm to others. We minimize the presence of harmful content by incorporating an LLM-judge that is tasked with refusing particular generations. For the chatbot side, however, part of our goal is to detect the frequency of such issues in their generations. As such, we decided against conducting safety filtering on chatbot responses. However, we acknowledge that our benchmark is not comprehensive enough to evaluate specialised content safety classification models.

\section*{Acknowledgments}

We would like to thank Bruno Martins and Luis Fernando D'Haro for their helpful feedback. We also want to give thanks to MS Azure services and Irving Kwong for their sponsorship.
This research was supported by the Portuguese Recovery and Resilience Plan through project C645008882-00000055 (Responsible.AI) and by national funds through \textit{Fundação para a Ciência e a Tecnologia, I.P.} (FCT) with references PRT/BD/152198/2021 (DOI: \url{doi.org/10.54499/PRT/BD/152198/2021}), UID/50021/2025 (DOI: \url{doi.org/10.54499/UID/50021/2025}) and UID/PRR/50021/2025 (DOI: \url{doi.org/10.54499/UID/PRR/50021/2025}).

\bibliography{anthology_0,anthology_1,custom}

@misc{zhang2021automatic,
      title={Automatic Evaluation and Moderation of Open-domain Dialogue Systems}, 
      author={Chen Zhang and João Sedoc and Luis Fernando D'Haro and Rafael Banchs and Alexander Rudnicky},
      year={2021},
      eprint={2111.02110},
      archivePrefix={arXiv},
      primaryClass={cs.CL}
}

@article{brown2020language,
  title={Language models are few-shot learners},
  author={Brown, Tom and Mann, Benjamin and Ryder, Nick and Subbiah, Melanie and Kaplan, Jared D and Dhariwal, Prafulla and Neelakantan, Arvind and Shyam, Pranav and Sastry, Girish and Askell, Amanda and others},
  journal={Advances in neural information processing systems},
  volume={33},
  pages={1877--1901},
  year={2020}
}

@article{Shuster2022BlenderBot3,
  title={BlenderBot 3: a deployed conversational agent that continually learns to responsibly engage},
  author={Kurt Shuster and Jing Xu and Mojtaba Komeili and Da Ju and Eric Michael Smith and Stephen Roller and Megan Ung and Moya Chen and Kushal Arora and Joshua Lane and Morteza Behrooz and W.K.F. Ngan and Spencer Poff and Naman Goyal and Arthur D. Szlam and Y-Lan Boureau and Melanie Kambadur and Jason Weston},
  journal={ArXiv},
  year={2022},
  volume={abs/2208.03188}
}

@misc{ouyang2022training,
      title={Training language models to follow instructions with human feedback}, 
      author={Long Ouyang and Jeff Wu and Xu Jiang and Diogo Almeida and Carroll L. Wainwright and Pamela Mishkin and Chong Zhang and Sandhini Agarwal and Katarina Slama and Alex Ray and John Schulman and Jacob Hilton and Fraser Kelton and Luke Miller and Maddie Simens and Amanda Askell and Peter Welinder and Paul Christiano and Jan Leike and Ryan Lowe},
      year={2022},
      eprint={2203.02155},
      archivePrefix={arXiv},
      primaryClass={cs.CL}
}

@article{DBLP:journals/corr/abs-2001-09977,
  author       = {Daniel Adiwardana and
                  Minh{-}Thang Luong and
                  David R. So and
                  Jamie Hall and
                  Noah Fiedel and
                  Romal Thoppilan and
                  Zi Yang and
                  Apoorv Kulshreshtha and
                  Gaurav Nemade and
                  Yifeng Lu and
                  Quoc V. Le},
  title        = {{Towards a Human-like Open-Domain Chatbot}},
  journal      = {CoRR},
  volume       = {abs/2001.09977},
  year         = {2020},
  url          = {https://arxiv.org/abs/2001.09977},
  eprinttype    = {arXiv},
  eprint       = {2001.09977},
  bibsource    = {dblp computer science bibliography, https://dblp.org}
}

@inproceedings{10.5555/2969033.2969173, author = {Sutskever, Ilya and Vinyals, Oriol and Le, Quoc V.}, title = {Sequence to sequence learning with neural networks}, year = {2014}, publisher = {MIT Press}, address = {Cambridge, MA, USA}, booktitle = {Proceedings of the 27th International Conference on Neural Information Processing Systems - Volume 2}, pages = {3104–3112}, numpages = {9}, location = {Montreal, Canada}, series = {NIPS'14} }

@inproceedings{10.5555/3016387.3016435, author = {Serban, Iulian V. and Sordoni, Alessandro and Bengio, Yoshua and Courville, Aaron and Pineau, Joelle}, title = {{Building end-to-end dialogue systems using generative hierarchical neural network models}}, year = {2016}, publisher = {AAAI Press}, booktitle = {Proceedings of the Thirtieth AAAI Conference on Artificial Intelligence}, pages = {3776–3783}, numpages = {8}, location = {Phoenix, Arizona}, series = {AAAI'16} }

@misc{openai2024gpt4,
      title={{GPT-4 Technical Report}}, 
      author={OpenAI},
      year={2024},
      eprint={2303.08774},
      archivePrefix={arXiv},
      primaryClass={cs.CL}
}

@Inproceedings{Gopalakrishnan2019,
 author = {Karthik Gopalakrishnan and Behnam Hedayatnia and Qinlang Chen and Anna Gottardi and Sanjeev Kwatra and Anushree Venkatesh and Raefer Gabriel and Dilek Hakkani-Tür},
 title = {Topical-Chat: Towards knowledge-grounded open-domain conversations},
 year = {2019},
 url = {https://www.amazon.science/publications/topical-chat-towards-knowledge-grounded-open-domain-conversations},
 booktitle = {Interspeech 2019},
}

@article{10.5555/3455716.3455856,
author = {Raffel, Colin and Shazeer, Noam and Roberts, Adam and Lee, Katherine and Narang, Sharan and Matena, Michael and Zhou, Yanqi and Li, Wei and Liu, Peter J.},
title = {Exploring the limits of transfer learning with a unified text-to-text transformer},
year = {2020},
issue_date = {January 2020},
publisher = {JMLR.org},
volume = {21},
number = {1},
issn = {1532-4435},
journal = {J. Mach. Learn. Res.},
month = {jan},
articleno = {140},
numpages = {67}
}

@inproceedings{46201,
author = {Vaswani, Ashish and Shazeer, Noam and Parmar, Niki and Uszkoreit, Jakob and Jones, Llion and Gomez, Aidan N. and Kaiser, \L{}ukasz and Polosukhin, Illia},
title = {Attention is all you need},
year = {2017},
isbn = {9781510860964},
publisher = {Curran Associates Inc.},
address = {Red Hook, NY, USA},
booktitle = {Proceedings of the 31st International Conference on Neural Information Processing Systems},
pages = {6000–6010},
numpages = {11},
location = {Long Beach, California, USA},
series = {NIPS'17}
}

@article{10.1162/neco.1997.9.8.1735,
author = {Hochreiter, Sepp and Schmidhuber, J\"{u}rgen},
title = {Long Short-Term Memory},
year = {1997},
issue_date = {November 15, 1997},
publisher = {MIT Press},
address = {Cambridge, MA, USA},
volume = {9},
number = {8},
issn = {0899-7667},
url = {https://doi.org/10.1162/neco.1997.9.8.1735},
doi = {10.1162/neco.1997.9.8.1735},
journal = {Neural Comput.},
month = {nov},
pages = {1735–1780},
numpages = {46}
}

@misc{perez2022discovering,
      title={Discovering Language Model Behaviors with Model-Written Evaluations}, 
      author={Ethan Perez and Sam Ringer and Kamilė Lukošiūtė and Karina Nguyen and Edwin Chen and Scott Heiner and Craig Pettit and Catherine Olsson and Sandipan Kundu and Saurav Kadavath and Andy Jones and Anna Chen and Ben Mann and Brian Israel and Bryan Seethor and Cameron McKinnon and Christopher Olah and Da Yan and Daniela Amodei and Dario Amodei and Dawn Drain and Dustin Li and Eli Tran-Johnson and Guro Khundadze and Jackson Kernion and James Landis and Jamie Kerr and Jared Mueller and Jeeyoon Hyun and Joshua Landau and Kamal Ndousse and Landon Goldberg and Liane Lovitt and Martin Lucas and Michael Sellitto and Miranda Zhang and Neerav Kingsland and Nelson Elhage and Nicholas Joseph and Noemí Mercado and Nova DasSarma and Oliver Rausch and Robin Larson and Sam McCandlish and Scott Johnston and Shauna Kravec and Sheer El Showk and Tamera Lanham and Timothy Telleen-Lawton and Tom Brown and Tom Henighan and Tristan Hume and Yuntao Bai and Zac Hatfield-Dodds and Jack Clark and Samuel R. Bowman and Amanda Askell and Roger Grosse and Danny Hernandez and Deep Ganguli and Evan Hubinger and Nicholas Schiefer and Jared Kaplan},
      year={2022},
      eprint={2212.09251},
      archivePrefix={arXiv},
      primaryClass={cs.CL},}

@misc{abdin2024phi3,
      title={Phi-3 Technical Report: A Highly Capable Language Model Locally on Your Phone}, 
      author={Marah Abdin and Sam Ade Jacobs and Ammar Ahmad Awan and Jyoti Aneja and Ahmed Awadallah and Hany Awadalla and Nguyen Bach and Amit Bahree and Arash Bakhtiari and Jianmin Bao and Harkirat Behl and Alon Benhaim and Misha Bilenko and Johan Bjorck and Sébastien Bubeck and Qin Cai and Martin Cai and Caio César Teodoro Mendes and Weizhu Chen and Vishrav Chaudhary and Dong Chen and Dongdong Chen and Yen-Chun Chen and Yi-Ling Chen and Parul Chopra and Xiyang Dai and Allie Del Giorno and Gustavo de Rosa and Matthew Dixon and Ronen Eldan and Victor Fragoso and Dan Iter and Mei Gao and Min Gao and Jianfeng Gao and Amit Garg and Abhishek Goswami and Suriya Gunasekar and Emman Haider and Junheng Hao and Russell J. Hewett and Jamie Huynh and Mojan Javaheripi and Xin Jin and Piero Kauffmann and Nikos Karampatziakis and Dongwoo Kim and Mahoud Khademi and Lev Kurilenko and James R. Lee and Yin Tat Lee and Yuanzhi Li and Yunsheng Li and Chen Liang and Lars Liden and Ce Liu and Mengchen Liu and Weishung Liu and Eric Lin and Zeqi Lin and Chong Luo and Piyush Madan and Matt Mazzola and Arindam Mitra and Hardik Modi and Anh Nguyen and Brandon Norick and Barun Patra and Daniel Perez-Becker and Thomas Portet and Reid Pryzant and Heyang Qin and Marko Radmilac and Corby Rosset and Sambudha Roy and Olatunji Ruwase and Olli Saarikivi and Amin Saied and Adil Salim and Michael Santacroce and Shital Shah and Ning Shang and Hiteshi Sharma and Swadheen Shukla and Xia Song and Masahiro Tanaka and Andrea Tupini and Xin Wang and Lijuan Wang and Chunyu Wang and Yu Wang and Rachel Ward and Guanhua Wang and Philipp Witte and Haiping Wu and Michael Wyatt and Bin Xiao and Can Xu and Jiahang Xu and Weijian Xu and Sonali Yadav and Fan Yang and Jianwei Yang and Ziyi Yang and Yifan Yang and Donghan Yu and Lu Yuan and Chengruidong Zhang and Cyril Zhang and Jianwen Zhang and Li Lyna Zhang and Yi Zhang and Yue Zhang and Yunan Zhang and Xiren Zhou},
      year={2024},
      eprint={2404.14219},
      archivePrefix={arXiv},
      primaryClass={cs.CL}
}

@article{russell1980circumplex,
  title={A circumplex model of affect},
  author={Russell, James A},
  journal={Journal of personality and social psychology},
  volume={39},
  number={6},
  pages={1161},
  year={1980},
  publisher={American Psychological Association}
}

@misc{ge2024scalingsyntheticdatacreation,
      title={{Scaling Synthetic Data Creation with 1,000,000,000 Personas}}, 
      author={Tao Ge and Xin Chan and Xiaoyang Wang and Dian Yu and Haitao Mi and Dong Yu},
      year={2024},
      eprint={2406.20094},
      archivePrefix={arXiv},
      primaryClass={cs.CL},
      url={https://arxiv.org/abs/2406.20094}, 
}

@misc{ibrahim2025multiturnevaluationanthropomorphicbehaviours,
      title={{Multi-turn Evaluation of Anthropomorphic Behaviours in Large Language Models}}, 
      author={Lujain Ibrahim and Canfer Akbulut and Rasmi Elasmar and Charvi Rastogi and Minsuk Kahng and Meredith Ringel Morris and Kevin R. McKee and Verena Rieser and Murray Shanahan and Laura Weidinger},
      year={2025},
      eprint={2502.07077},
      archivePrefix={arXiv},
      primaryClass={cs.CL},
      url={https://arxiv.org/abs/2502.07077}, 
}

@inproceedings{ma-etal-2025-communication,
    title = "Communication Makes Perfect: Persuasion Dataset Construction via Multi-{LLM} Communication",
    author = "Ma, Weicheng  and
      Zhang, Hefan  and
      Yang, Ivory  and
      Ji, Shiyu  and
      Chen, Joice  and
      Hashemi, Farnoosh  and
      Mohole, Shubham  and
      Gearey, Ethan  and
      Macy, Michael  and
      Hassanpour, Saeed  and
      Vosoughi, Soroush",
    editor = "Chiruzzo, Luis  and
      Ritter, Alan  and
      Wang, Lu",
    booktitle = "Proceedings of the 2025 Conference of the Nations of the Americas Chapter of the Association for Computational Linguistics: Human Language Technologies (Volume 1: Long Papers)",
    month = apr,
    year = "2025",
    address = "Albuquerque, New Mexico",
    publisher = "Association for Computational Linguistics",
    url = "https://aclanthology.org/2025.naacl-long.203/",
    pages = "4017--4045",
    ISBN = "979-8-89176-189-6"
}

@misc{guo2024largelanguagemodelbased,
      title={{Large Language Model based Multi-Agents: A Survey of Progress and Challenges}}, 
      author={Taicheng Guo and Xiuying Chen and Yaqi Wang and Ruidi Chang and Shichao Pei and Nitesh V. Chawla and Olaf Wiest and Xiangliang Zhang},
      year={2024},
      eprint={2402.01680},
      archivePrefix={arXiv},
      primaryClass={cs.CL},
      url={https://arxiv.org/abs/2402.01680}, 
}

@misc{lin2024diversedialoguemethodologydesigningchatbots,
      title={{DiverseDialogue: A Methodology for Designing Chatbots with Human-Like Diversity}}, 
      author={Xiaoyu Lin and Xinkai Yu and Ankit Aich and Salvatore Giorgi and Lyle Ungar},
      year={2024},
      eprint={2409.00262},
      archivePrefix={arXiv},
      primaryClass={cs.CL},
      url={https://arxiv.org/abs/2409.00262}, 
}

@inproceedings{
li2024culturegen,
title={{CULTURE}-{GEN}: Revealing Global Cultural Perception in Language Models through Natural Language Prompting},
author={Huihan Li and Liwei Jiang and Nouha Dziri and Xiang Ren and Yejin Choi},
booktitle={First Conference on Language Modeling},
year={2024},
url={https://openreview.net/forum?id=DbsLm2KAqP}
}

@inproceedings{
zheng2023judging,
title={Judging {LLM}-as-a-Judge with {MT}-Bench and Chatbot Arena},
author={Lianmin Zheng and Wei-Lin Chiang and Ying Sheng and Siyuan Zhuang and Zhanghao Wu and Yonghao Zhuang and Zi Lin and Zhuohan Li and Dacheng Li and Eric Xing and Hao Zhang and Joseph E. Gonzalez and Ion Stoica},
booktitle={Thirty-seventh Conference on Neural Information Processing Systems Datasets and Benchmarks Track},
year={2023},
url={https://openreview.net/forum?id=uccHPGDlao}
}

@misc{finch2020emorainquisitivesocialchatbot,
      title={{Emora: An Inquisitive Social Chatbot Who Cares For You}}, 
      author={Sarah E. Finch and James D. Finch and Ali Ahmadvand and Ingyu and Choi and Xiangjue Dong and Ruixiang Qi and Harshita Sahijwani and Sergey Volokhin and Zihan Wang and Zihao Wang and Jinho D. Choi},
      year={2020},
      eprint={2009.04617},
      archivePrefix={arXiv},
      primaryClass={cs.CL},
      url={https://arxiv.org/abs/2009.04617}, 
}

@article{10.1162/coli_a_00368,
    author = {Zhou, Li and Gao, Jianfeng and Li, Di and Shum, Heung-Yeung},
    title = {T{he Design and Implementation of XiaoIce, an Empathetic Social
                    Chatbot}},
    journal = {Computational Linguistics},
    volume = {46},
    number = {1},
    pages = {53-93},
    year = {2020},
    month = {03},
    issn = {0891-2017},
    doi = {10.1162/coli_a_00368},
    url = {https://doi.org/10.1162/coli\_a\_00368},
    eprint = {https://direct.mit.edu/coli/article-pdf/46/1/53/1847834/coli\_a\_00368.pdf},
}

@misc{gemmateam2025gemma3technicalreport,
      title={{Gemma 3 Technical Report}}, 
      author={Team Gemma},
      year={2025},
      eprint={2503.19786},
      archivePrefix={arXiv},
      primaryClass={cs.CL},
      url={https://arxiv.org/abs/2503.19786}, 
}

@misc{gpt4omini,
      title={{GPT-4o System Card}}, 
      author={OpenAI},
      year={2024},
      eprint={2410.21276},
      archivePrefix={arXiv},
      primaryClass={cs.CL},
      url={https://arxiv.org/abs/2410.21276}, 
}

@misc{dang2024ayaexpansecombiningresearch,
      title={{Aya Expanse: Combining Research Breakthroughs for a New Multilingual Frontier}}, 
      author={John Dang and Shivalika Singh and Daniel D'souza and Arash Ahmadian and Alejandro Salamanca and Madeline Smith and Aidan Peppin and Sungjin Hong and Manoj Govindassamy and Terrence Zhao and Sandra Kublik and Meor Amer and Viraat Aryabumi and Jon Ander Campos and Yi-Chern Tan and Tom Kocmi and Florian Strub and Nathan Grinsztajn and Yannis Flet-Berliac and Acyr Locatelli and Hangyu Lin and Dwarak Talupuru and Bharat Venkitesh and David Cairuz and Bowen Yang and Tim Chung and Wei-Yin Ko and Sylvie Shang Shi and Amir Shukayev and Sammie Bae and Aleksandra Piktus and Roman Castagné and Felipe Cruz-Salinas and Eddie Kim and Lucas Crawhall-Stein and Adrien Morisot and Sudip Roy and Phil Blunsom and Ivan Zhang and Aidan Gomez and Nick Frosst and Marzieh Fadaee and Beyza Ermis and Ahmet Üstün and Sara Hooker},
      year={2024},
      eprint={2412.04261},
      archivePrefix={arXiv},
      primaryClass={cs.CL},
      url={https://arxiv.org/abs/2412.04261}, 
}

@misc{grattafiori2024llama3herdmodels,
      title={{The Llama 3 Herd of Models}}, 
      author={Aaron Grattafiori and Abhimanyu Dubey and Abhinav Jauhri and Abhinav Pandey and Abhishek Kadian and Ahmad Al-Dahle and Aiesha Letman and Akhil Mathur and Alan Schelten and Alex Vaughan and Amy Yang and Angela Fan and Anirudh Goyal and Anthony Hartshorn and Aobo Yang and Archi Mitra and Archie Sravankumar and Artem Korenev and Arthur Hinsvark and Arun Rao and Aston Zhang and Aurelien Rodriguez and Austen Gregerson and Ava Spataru and Baptiste Roziere and Bethany Biron and Binh Tang and Bobbie Chern and Charlotte Caucheteux and Chaya Nayak and Chloe Bi and Chris Marra and Chris McConnell and Christian Keller and Christophe Touret and Chunyang Wu and Corinne Wong and Cristian Canton Ferrer and Cyrus Nikolaidis and Damien Allonsius and Daniel Song and Danielle Pintz and Danny Livshits and Danny Wyatt and David Esiobu and Dhruv Choudhary and Dhruv Mahajan and Diego Garcia-Olano and Diego Perino and Dieuwke Hupkes and Egor Lakomkin and Ehab AlBadawy and Elina Lobanova and Emily Dinan and Eric Michael Smith and Filip Radenovic and Francisco Guzmán and Frank Zhang and Gabriel Synnaeve and Gabrielle Lee and Georgia Lewis Anderson and Govind Thattai and Graeme Nail and Gregoire Mialon and Guan Pang and Guillem Cucurell and Hailey Nguyen and Hannah Korevaar and Hu Xu and Hugo Touvron and Iliyan Zarov and Imanol Arrieta Ibarra and Isabel Kloumann and Ishan Misra and Ivan Evtimov and Jack Zhang and Jade Copet and Jaewon Lee and Jan Geffert and Jana Vranes and Jason Park and Jay Mahadeokar and Jeet Shah and Jelmer van der Linde and Jennifer Billock and Jenny Hong and Jenya Lee and Jeremy Fu and Jianfeng Chi and Jianyu Huang and Jiawen Liu and Jie Wang and Jiecao Yu and Joanna Bitton and Joe Spisak and Jongsoo Park and Joseph Rocca and Joshua Johnstun and Joshua Saxe and Junteng Jia and Kalyan Vasuden Alwala and Karthik Prasad and Kartikeya Upasani and Kate Plawiak and Ke Li and Kenneth Heafield and Kevin Stone and Khalid El-Arini and Krithika Iyer and Kshitiz Malik and Kuenley Chiu and Kunal Bhalla and Kushal Lakhotia and Lauren Rantala-Yeary and Laurens van der Maaten and Lawrence Chen and Liang Tan and Liz Jenkins and Louis Martin and Lovish Madaan and Lubo Malo and Lukas Blecher and Lukas Landzaat and Luke de Oliveira and Madeline Muzzi and Mahesh Pasupuleti and Mannat Singh and Manohar Paluri and Marcin Kardas and Maria Tsimpoukelli and Mathew Oldham and Mathieu Rita and Maya Pavlova and Melanie Kambadur and Mike Lewis and Min Si and Mitesh Kumar Singh and Mona Hassan and Naman Goyal and Narjes Torabi and Nikolay Bashlykov and Nikolay Bogoychev and Niladri Chatterji and Ning Zhang and Olivier Duchenne and Onur Çelebi and Patrick Alrassy and Pengchuan Zhang and Pengwei Li and Petar Vasic and Peter Weng and Prajjwal Bhargava and Pratik Dubal and Praveen Krishnan and Punit Singh Koura and Puxin Xu and Qing He and Qingxiao Dong and Ragavan Srinivasan and Raj Ganapathy and Ramon Calderer and Ricardo Silveira Cabral and Robert Stojnic and Roberta Raileanu and Rohan Maheswari and Rohit Girdhar and Rohit Patel and Romain Sauvestre and Ronnie Polidoro and Roshan Sumbaly and Ross Taylor and Ruan Silva and Rui Hou and Rui Wang and Saghar Hosseini and Sahana Chennabasappa and Sanjay Singh and Sean Bell and Seohyun Sonia Kim and Sergey Edunov and Shaoliang Nie and Sharan Narang and Sharath Raparthy and Sheng Shen and Shengye Wan and Shruti Bhosale and Shun Zhang and Simon Vandenhende and Soumya Batra and Spencer Whitman and Sten Sootla and Stephane Collot and Suchin Gururangan and Sydney Borodinsky and Tamar Herman and Tara Fowler and Tarek Sheasha and Thomas Georgiou and Thomas Scialom and Tobias Speckbacher and Todor Mihaylov and Tong Xiao and Ujjwal Karn and Vedanuj Goswami and Vibhor Gupta and Vignesh Ramanathan and Viktor Kerkez and Vincent Gonguet and Virginie Do and Vish Vogeti and Vítor Albiero and Vladan Petrovic and Weiwei Chu and Wenhan Xiong and Wenyin Fu and Whitney Meers and Xavier Martinet and Xiaodong Wang and Xiaofang Wang and Xiaoqing Ellen Tan and Xide Xia and Xinfeng Xie and Xuchao Jia and Xuewei Wang and Yaelle Goldschlag and Yashesh Gaur and Yasmine Babaei and Yi Wen and Yiwen Song and Yuchen Zhang and Yue Li and Yuning Mao and Zacharie Delpierre Coudert and Zheng Yan and Zhengxing Chen and Zoe Papakipos and Aaditya Singh and Aayushi Srivastava and Abha Jain and Adam Kelsey and Adam Shajnfeld and Adithya Gangidi and Adolfo Victoria and Ahuva Goldstand and Ajay Menon and Ajay Sharma and Alex Boesenberg and Alexei Baevski and Allie Feinstein and Amanda Kallet and Amit Sangani and Amos Teo and Anam Yunus and Andrei Lupu and Andres Alvarado and Andrew Caples and Andrew Gu and Andrew Ho and Andrew Poulton and Andrew Ryan and Ankit Ramchandani and Annie Dong and Annie Franco and Anuj Goyal and Aparajita Saraf and Arkabandhu Chowdhury and Ashley Gabriel and Ashwin Bharambe and Assaf Eisenman and Azadeh Yazdan and Beau James and Ben Maurer and Benjamin Leonhardi and Bernie Huang and Beth Loyd and Beto De Paola and Bhargavi Paranjape and Bing Liu and Bo Wu and Boyu Ni and Braden Hancock and Bram Wasti and Brandon Spence and Brani Stojkovic and Brian Gamido and Britt Montalvo and Carl Parker and Carly Burton and Catalina Mejia and Ce Liu and Changhan Wang and Changkyu Kim and Chao Zhou and Chester Hu and Ching-Hsiang Chu and Chris Cai and Chris Tindal and Christoph Feichtenhofer and Cynthia Gao and Damon Civin and Dana Beaty and Daniel Kreymer and Daniel Li and David Adkins and David Xu and Davide Testuggine and Delia David and Devi Parikh and Diana Liskovich and Didem Foss and Dingkang Wang and Duc Le and Dustin Holland and Edward Dowling and Eissa Jamil and Elaine Montgomery and Eleonora Presani and Emily Hahn and Emily Wood and Eric-Tuan Le and Erik Brinkman and Esteban Arcaute and Evan Dunbar and Evan Smothers and Fei Sun and Felix Kreuk and Feng Tian and Filippos Kokkinos and Firat Ozgenel and Francesco Caggioni and Frank Kanayet and Frank Seide and Gabriela Medina Florez and Gabriella Schwarz and Gada Badeer and Georgia Swee and Gil Halpern and Grant Herman and Grigory Sizov and Guangyi and Zhang and Guna Lakshminarayanan and Hakan Inan and Hamid Shojanazeri and Han Zou and Hannah Wang and Hanwen Zha and Haroun Habeeb and Harrison Rudolph and Helen Suk and Henry Aspegren and Hunter Goldman and Hongyuan Zhan and Ibrahim Damlaj and Igor Molybog and Igor Tufanov and Ilias Leontiadis and Irina-Elena Veliche and Itai Gat and Jake Weissman and James Geboski and James Kohli and Janice Lam and Japhet Asher and Jean-Baptiste Gaya and Jeff Marcus and Jeff Tang and Jennifer Chan and Jenny Zhen and Jeremy Reizenstein and Jeremy Teboul and Jessica Zhong and Jian Jin and Jingyi Yang and Joe Cummings and Jon Carvill and Jon Shepard and Jonathan McPhie and Jonathan Torres and Josh Ginsburg and Junjie Wang and Kai Wu and Kam Hou U and Karan Saxena and Kartikay Khandelwal and Katayoun Zand and Kathy Matosich and Kaushik Veeraraghavan and Kelly Michelena and Keqian Li and Kiran Jagadeesh and Kun Huang and Kunal Chawla and Kyle Huang and Lailin Chen and Lakshya Garg and Lavender A and Leandro Silva and Lee Bell and Lei Zhang and Liangpeng Guo and Licheng Yu and Liron Moshkovich and Luca Wehrstedt and Madian Khabsa and Manav Avalani and Manish Bhatt and Martynas Mankus and Matan Hasson and Matthew Lennie and Matthias Reso and Maxim Groshev and Maxim Naumov and Maya Lathi and Meghan Keneally and Miao Liu and Michael L. Seltzer and Michal Valko and Michelle Restrepo and Mihir Patel and Mik Vyatskov and Mikayel Samvelyan and Mike Clark and Mike Macey and Mike Wang and Miquel Jubert Hermoso and Mo Metanat and Mohammad Rastegari and Munish Bansal and Nandhini Santhanam and Natascha Parks and Natasha White and Navyata Bawa and Nayan Singhal and Nick Egebo and Nicolas Usunier and Nikhil Mehta and Nikolay Pavlovich Laptev and Ning Dong and Norman Cheng and Oleg Chernoguz and Olivia Hart and Omkar Salpekar and Ozlem Kalinli and Parkin Kent and Parth Parekh and Paul Saab and Pavan Balaji and Pedro Rittner and Philip Bontrager and Pierre Roux and Piotr Dollar and Polina Zvyagina and Prashant Ratanchandani and Pritish Yuvraj and Qian Liang and Rachad Alao and Rachel Rodriguez and Rafi Ayub and Raghotham Murthy and Raghu Nayani and Rahul Mitra and Rangaprabhu Parthasarathy and Raymond Li and Rebekkah Hogan and Robin Battey and Rocky Wang and Russ Howes and Ruty Rinott and Sachin Mehta and Sachin Siby and Sai Jayesh Bondu and Samyak Datta and Sara Chugh and Sara Hunt and Sargun Dhillon and Sasha Sidorov and Satadru Pan and Saurabh Mahajan and Saurabh Verma and Seiji Yamamoto and Sharadh Ramaswamy and Shaun Lindsay and Shaun Lindsay and Sheng Feng and Shenghao Lin and Shengxin Cindy Zha and Shishir Patil and Shiva Shankar and Shuqiang Zhang and Shuqiang Zhang and Sinong Wang and Sneha Agarwal and Soji Sajuyigbe and Soumith Chintala and Stephanie Max and Stephen Chen and Steve Kehoe and Steve Satterfield and Sudarshan Govindaprasad and Sumit Gupta and Summer Deng and Sungmin Cho and Sunny Virk and Suraj Subramanian and Sy Choudhury and Sydney Goldman and Tal Remez and Tamar Glaser and Tamara Best and Thilo Koehler and Thomas Robinson and Tianhe Li and Tianjun Zhang and Tim Matthews and Timothy Chou and Tzook Shaked and Varun Vontimitta and Victoria Ajayi and Victoria Montanez and Vijai Mohan and Vinay Satish Kumar and Vishal Mangla and Vlad Ionescu and Vlad Poenaru and Vlad Tiberiu Mihailescu and Vladimir Ivanov and Wei Li and Wenchen Wang and Wenwen Jiang and Wes Bouaziz and Will Constable and Xiaocheng Tang and Xiaojian Wu and Xiaolan Wang and Xilun Wu and Xinbo Gao and Yaniv Kleinman and Yanjun Chen and Ye Hu and Ye Jia and Ye Qi and Yenda Li and Yilin Zhang and Ying Zhang and Yossi Adi and Youngjin Nam and Yu and Wang and Yu Zhao and Yuchen Hao and Yundi Qian and Yunlu Li and Yuzi He and Zach Rait and Zachary DeVito and Zef Rosnbrick and Zhaoduo Wen and Zhenyu Yang and Zhiwei Zhao and Zhiyu Ma},
      year={2024},
      eprint={2407.21783},
      archivePrefix={arXiv},
      primaryClass={cs.AI},
      url={https://arxiv.org/abs/2407.21783}, 
}

@misc{Google_2025,
    url={https://blog.google/technology/google-deepmind/google-gemini-ai-update-december-2024/#ceo-message},
    title={{Introducing Gemini 2.0: our new AI model for the agentic era}},
    author={Team Gemini and Google},
    year={2025}
}

@misc{saito2023verbositybiaspreferencelabeling,
      title={{Verbosity Bias in Preference Labeling by Large Language Models}}, 
      author={Keita Saito and Akifumi Wachi and Koki Wataoka and Youhei Akimoto},
      year={2023},
      eprint={2310.10076},
      archivePrefix={arXiv},
      primaryClass={cs.CL},
      url={https://arxiv.org/abs/2310.10076}, 
}

@Article{Qian2025,
 author = {Yiyue Qian and Shinan Zhang and Yun Zhou and Haibo Ding and Diego Socolinsky and Yi Zhang},
 title = {{Enhancing LLM-as-a-judge via multi-agent collaboration}},
 year = {2025},
 url = {https://www.amazon.science/publications/enhancing-llm-as-a-judge-via-multi-agent-collaboration},
}

@inproceedings{wu-aji-2025-style,
    title = "Style Over Substance: Evaluation Biases for Large Language Models",
    author = "Wu, Minghao  and
      Aji, Alham Fikri",
    editor = "Rambow, Owen  and
      Wanner, Leo  and
      Apidianaki, Marianna  and
      Al-Khalifa, Hend  and
      Eugenio, Barbara Di  and
      Schockaert, Steven",
    booktitle = "Proceedings of the 31st International Conference on Computational Linguistics",
    month = jan,
    year = "2025",
    address = "Abu Dhabi, UAE",
    publisher = "Association for Computational Linguistics",
    url = "https://aclanthology.org/2025.coling-main.21/",
    pages = "297--312"
}

@InProceedings{pmlr-v202-aher23a,
  title = 	 {{Using Large Language Models to Simulate Multiple Humans and Replicate Human Subject Studies}},
  author =       {Aher, Gati V and Arriaga, Rosa I. and Kalai, Adam Tauman},
  booktitle = 	 {Proceedings of the 40th International Conference on Machine Learning},
  pages = 	 {337--371},
  year = 	 {2023},
  editor = 	 {Krause, Andreas and Brunskill, Emma and Cho, Kyunghyun and Engelhardt, Barbara and Sabato, Sivan and Scarlett, Jonathan},
  volume = 	 {202},
  series = 	 {Proceedings of Machine Learning Research},
  month = 	 {23--29 Jul},
  publisher =    {PMLR},
  url = 	 {https://proceedings.mlr.press/v202/aher23a.html}
}

@misc{huang2024conceptevaluationprotocol,
      title={{Concept -- An Evaluation Protocol on Conversational Recommender Systems with System-centric and User-centric Factors}}, 
      author={Chen Huang and Peixin Qin and Yang Deng and Wenqiang Lei and Jiancheng Lv and Tat-Seng Chua},
      year={2024},
      eprint={2404.03304},
      archivePrefix={arXiv},
      primaryClass={cs.CL},
      url={https://arxiv.org/abs/2404.03304}, 
}

@InProceedings{pmlr-v139-kossen21a,
  title = 	 {{Active Testing: Sample-Efficient Model Evaluation}},
  author =       {Kossen, Jannik and Farquhar, Sebastian and Gal, Yarin and Rainforth, Tom},
  booktitle = 	 {Proceedings of the 38th International Conference on Machine Learning},
  pages = 	 {5753--5763},
  year = 	 {2021},
  editor = 	 {Meila, Marina and Zhang, Tong},
  volume = 	 {139},
  series = 	 {Proceedings of Machine Learning Research},
  month = 	 {18--24 Jul},
  publisher =    {PMLR},
  url = 	 {https://proceedings.mlr.press/v139/kossen21a.html}
}

@misc{li2024activeevaluationacquisitionefficient,
      title={{Active Evaluation Acquisition for Efficient LLM Benchmarking}}, 
      author={Yang Li and Jie Ma and Miguel Ballesteros and Yassine Benajiba and Graham Horwood},
      year={2024},
      eprint={2410.05952},
      archivePrefix={arXiv},
      primaryClass={cs.LG},
      url={https://arxiv.org/abs/2410.05952}, 
}

@misc{o3,
      title={{OpenAI o3 and o4-mini System Card}}, 
      author={OpenAI},
      year={2025},
      url={https://openai.com/index/o3-o4-mini-system-card/}
}

@misc{deepseekai2025deepseekv3technicalreport,
      title={{DeepSeek-V3 Technical Report}}, 
      author={DeepSeek-AI},
      year={2025},
      eprint={2412.19437},
      archivePrefix={arXiv},
      primaryClass={cs.CL},
      url={https://arxiv.org/abs/2412.19437}, 
}

@misc{deepseekai2025deepseekr1incentivizingreasoningcapability,
      title={{DeepSeek-R1: Incentivizing Reasoning Capability in LLMs via Reinforcement Learning}}, 
      author={DeepSeek-AI},
      year={2025},
      eprint={2501.12948},
      archivePrefix={arXiv},
      primaryClass={cs.CL},
      url={https://arxiv.org/abs/2501.12948}, 
}

@article{mtld,
	author = {McCarthy, Philip M. and Jarvis, Scott},
	date = {2010/05/01},
	doi = {10.3758/BRM.42.2.381},
	id = {McCarthy2010},
	isbn = {1554-3528},
	journal = {Behavior Research Methods},
	number = {2},
	pages = {381--392},
	title = {MTLD, vocd-D, and HD-D: A validation study of sophisticated approaches to lexical diversity assessment},
	url = {https://doi.org/10.3758/BRM.42.2.381},
	volume = {42},
	year = {2010}}

@misc{liu2025mdsevalmetaevaluationbenchmarkmultimodal,
      title={MDSEval: A Meta-Evaluation Benchmark for Multimodal Dialogue Summarization}, 
      author={Yinhong Liu and Jianfeng He and Hang Su and Ruixue Lian and Yi Nian and Jake Vincent and Srikanth Vishnubhotla and Robinson Piramuthu and Saab Mansour},
      year={2025},
      eprint={2510.01659},
      archivePrefix={arXiv},
      primaryClass={cs.CL},
      url={https://arxiv.org/abs/2510.01659}, 
}

@inproceedings{
yue2025pangea,
title={Pangea: A Fully Open Multilingual Multimodal {LLM} for 39 Languages},
author={Xiang Yue and Yueqi Song and Akari Asai and Seungone Kim and Jean de Dieu Nyandwi and Simran Khanuja and Anjali Kantharuban and Lintang Sutawika and Sathyanarayanan Ramamoorthy and Graham Neubig},
booktitle={The Thirteenth International Conference on Learning Representations},
year={2025},
url={https://openreview.net/forum?id=a3g2l4yEys}
}

@misc{qwen3technicalreport,
      title={{Qwen3 Technical Report}}, 
      author={Team Qwen},
      year={2025},
      eprint={2505.09388},
      archivePrefix={arXiv},
      primaryClass={cs.CL},
      url={https://arxiv.org/abs/2505.09388}, 
}

@misc{kimiteam2025kimik2openagentic,
      title={{Kimi K2: Open Agentic Intelligence}}, 
      author={Team Kimi},
      year={2025},
      eprint={2507.20534},
      archivePrefix={arXiv},
      primaryClass={cs.LG},
      url={https://arxiv.org/abs/2507.20534}, 
}

@misc{singh2025openaigpt5card,
      title={{OpenAI GPT-5 System Card}}, 
      author={OpenAI},
      year={2025},
      eprint={2601.03267},
      archivePrefix={arXiv},
      primaryClass={cs.CL},
      url={https://arxiv.org/abs/2601.03267}, 
}

@inproceedings{zhang-etal-2025-p,
    title = "{P}-{MME}val: A Parallel Multilingual Multitask Benchmark for Consistent Evaluation of {LLM}s",
    author = "Zhang, Yidan  and
      Wan, Yu  and
      Deng, Boyi  and
      Yang, Baosong  and
      Wei, Hao-Ran  and
      Huang, Fei  and
      Yu, Bowen  and
      Liu, Dayiheng  and
      Lin, Junyang  and
      Huang, Fei  and
      Zhou, Jingren",
    editor = "Christodoulopoulos, Christos  and
      Chakraborty, Tanmoy  and
      Rose, Carolyn  and
      Peng, Violet",
    booktitle = "Proceedings of the 2025 Conference on Empirical Methods in Natural Language Processing",
    month = nov,
    year = "2025",
    address = "Suzhou, China",
    publisher = "Association for Computational Linguistics",
    url = "https://aclanthology.org/2025.emnlp-main.242/",
    doi = "10.18653/v1/2025.emnlp-main.242",
    pages = "4809--4836",
    ISBN = "979-8-89176-332-6"
}

@misc{qwen2025qwen25technicalreport,
      title={{Qwen2.5 Technical Report}}, 
      author={Team Qwen},
      year={2025},
      eprint={2412.15115},
      archivePrefix={arXiv},
      primaryClass={cs.CL},
      url={https://arxiv.org/abs/2412.15115}, 
}

@inproceedings{
romanou2025include,
title={{INCLUDE}: Evaluating Multilingual Language Understanding with Regional Knowledge},
author={Angelika Romanou and Negar Foroutan and Anna Sotnikova and Sree Harsha Nelaturu and Shivalika Singh and Rishabh Maheshwary and Micol Altomare and Zeming Chen and Mohamed A. Haggag and Snegha A and Alfonso Amayuelas and Azril Hafizi Amirudin and Danylo Boiko and Michael Chang and Jenny Chim and Gal Cohen and Aditya Kumar Dalmia and Abraham Diress and Sharad Duwal and Daniil Dzenhaliou and Daniel Fernando Erazo Florez and Fabian Farestam and Joseph Marvin Imperial and Shayekh Bin Islam and Perttu Isotalo and Maral Jabbarishiviari and B{\"o}rje F. Karlsson and Eldar Khalilov and Christopher Klamm and Fajri Koto and Dominik Krzemi{\'n}ski and Gabriel Adriano de Melo and Syrielle Montariol and Yiyang Nan and Joel Niklaus and Jekaterina Novikova and Johan Samir Obando Ceron and Debjit Paul and Esther Ploeger and Jebish Purbey and Swati Rajwal and Selvan Sunitha Ravi and Sara Rydell and Roshan Santhosh and Drishti Sharma and Marjana Prifti Skenduli and Arshia Soltani Moakhar and Bardia soltani moakhar and Ayush Kumar Tarun and Azmine Toushik Wasi and Thenuka Ovin Weerasinghe and Serhan Yilmaz and Mike Zhang and Imanol Schlag and Marzieh Fadaee and Sara Hooker and Antoine Bosselut},
booktitle={The Thirteenth International Conference on Learning Representations},
year={2025},
url={https://openreview.net/forum?id=k3gCieTXeY}
}

@inproceedings{singh-etal-2025-global,
    title = "Global {MMLU}: Understanding and Addressing Cultural and Linguistic Biases in Multilingual Evaluation",
    author = "Singh, Shivalika  and
      Romanou, Angelika  and
      Fourrier, Cl{\'e}mentine  and
      Adelani, David Ifeoluwa  and
      Ngui, Jian Gang  and
      Vila-Suero, Daniel  and
      Limkonchotiwat, Peerat  and
      Marchisio, Kelly  and
      Leong, Wei Qi  and
      Susanto, Yosephine  and
      Ng, Raymond  and
      Longpre, Shayne  and
      Ruder, Sebastian  and
      Ko, Wei-Yin  and
      Bosselut, Antoine  and
      Oh, Alice  and
      Martins, Andre  and
      Choshen, Leshem  and
      Ippolito, Daphne  and
      Ferrante, Enzo  and
      Fadaee, Marzieh  and
      Ermis, Beyza  and
      Hooker, Sara",
    editor = "Che, Wanxiang  and
      Nabende, Joyce  and
      Shutova, Ekaterina  and
      Pilehvar, Mohammad Taher",
    booktitle = "Proceedings of the 63rd Annual Meeting of the Association for Computational Linguistics (Volume 1: Long Papers)",
    month = jul,
    year = "2025",
    address = "Vienna, Austria",
    publisher = "Association for Computational Linguistics",
    url = "https://aclanthology.org/2025.acl-long.919/",
    doi = "10.18653/v1/2025.acl-long.919",
    pages = "18761--18799",
    ISBN = "979-8-89176-251-0"
}

@inproceedings{
yao2025taubench,
title={\{\${\textbackslash}tau\$\}-bench: A Benchmark for {\textbackslash}underline\{T\}ool-{\textbackslash}underline\{A\}gent-{\textbackslash}underline\{U\}ser Interaction in Real-World Domains},
author={Shunyu Yao and Noah Shinn and Pedram Razavi and Karthik R Narasimhan},
booktitle={The Thirteenth International Conference on Learning Representations},
year={2025},
url={https://openreview.net/forum?id=roNSXZpUDN}
}

@inproceedings{
hendrycks2021measuring,
title={Measuring Massive Multitask Language Understanding},
author={Dan Hendrycks and Collin Burns and Steven Basart and Andy Zou and Mantas Mazeika and Dawn Song and Jacob Steinhardt},
booktitle={International Conference on Learning Representations},
year={2021},
url={https://openreview.net/forum?id=d7KBjmI3GmQ}
}

\newpage

\appendix

\section{Meta-evaluation Benchmarks}
\label{sec:datasets}

This section presents a brief survey of the datasets that have been used as a benchmark for LLM-based open-domain dialogue evaluation metrics.

The \textbf{FED} dataset \citep{mehri-eskenazi-2020-unsupervised} contains conversations between humans and two chatbots, Meena \citep{DBLP:journals/corr/abs-2001-09977} and Mitsuku. It includes 124 dialogues (40 with Meena, 44 with Mitsuku, and 40 with humans), annotated for eighteen quality aspects. The dataset comprises 3,348 turn-level and 1,364 dialogue-level annotations, totalling 4,712 data points, with each conversation receiving one dialogue-level and three randomly selected turn-level annotations.

For the \textbf{USR} dataset \citep{mehri-eskenazi-2020-usr}, annotations were collected for responses generated by models trained on the TopicalChat \citep{Gopalakrishnan2019} and PersonaChat \citep{zhang-etal-2018-personalizing} datasets. The models included Transformer \citep{46201}, RNN Seq2Seq \citep{shang-etal-2015-neural}, LSTM \citep{10.1162/neco.1997.9.8.1735}, and KV-MemNN \citep{miller-etal-2016-key}. Annotations were performed on sixty dialogue contexts, each featuring multiple chatbot responses (six for Topical-Chat, five for PersonaChat) and an additional human response.

The \textbf{DSTC10} track \citep{zhang2021automatic} utilized a test set derived from three sources: CHANEL-JSALT2020, ChatEval \citep{sedoc-etal-2019-chateval}, and new annotations on TopicalChat \citep{Gopalakrishnan2019} and PersonaChat \citep{zhang-etal-2018-personalizing}. Responses were generated by eight systems --LSTM Seq2Seq, Attention-based LSTM Seq2Seq \citep{10.5555/2969033.2969173}, HRED \citep{10.5555/3016387.3016435}, VHRED, BlenderBot (400M-Distill) \citep{roller-etal-2021-recipes}, DialoGPT-medium \citep{zhang-etal-2020-dialogpt}, T5-base \citep{10.5555/3455716.3455856}, and GPT-3 \citep{brown2020language}—alongside a human baseline and random utterances.

The \textbf{DSTC11} test set \citep{rodriguez-cantelar-etal-2023-overview} combined a portion of the DSTC10 test set with new human-chatbot dialogues. These dialogues featured state-of-the-art chatbots: ChatGPT, GPT-3.5 \citep{ouyang2022training}, and BlenderBot3 \citep{Shuster2022BlenderBot3} for English; and Chinese DialoGPT, Xiaoice \citep{10.1162/coli_a_00368}, and Plato-XL \citep{bao-etal-2022-plato} for Chinese. The test set underwent translation and back-translation to English, Chinese, and Spanish. In total, it includes 4,839 turn-level and 277 dialogue-level annotations.

\textbf{XDial-Eval} \citep{zhang-etal-2023-xdial} is built on top of several existing English dialogue evaluation datasets, all of which predate the introduction of LLMs for dialogue. This curation process yields grand total of 14,930 annotated turns and 8,691 annotated dialogues. A commercial machine translation system is then used to extend these datasets to nine other languages: Chinese (ZH), Spanish (ES), German (DE), French (FR), Japanese (JA), Korean (KO), Hindi (HI), Arabic (AR), and Russian (RU).

In \textbf{ABC-Eval \citep{finch-etal-2023-dont}}, multi-dimensional annotations at the turn and dialogue level were collected on unique conversations in English between collaborative humans and 4 different chatbots: (1) \textbf{Blender-Decode} \citep{nie-etal-2021-like}; (2) \textbf{Blender2} \citep{xu-etal-2022-beyond}; (3) \textbf{BART-FiD-RAG} \citep{shuster-etal-2021-retrieval-augmentation}; and (4) \textbf{Emora} \citep{finch2020emorainquisitivesocialchatbot}. Annotations were obtained using ABC-Eval binary behaviour labels (16), dialogue-level (8), turn level (8) and comparative (8). In total, 400 dialogues were collected and annotated, 100 per chatbot.

For \textbf{Soda-Eval \citep{mendonca-etal-2024-soda}}, the authors obtain 120K turn level annotations stemming from 10K dialogues extracted from the Soda dataset \citep{kim-etal-2023-soda} which uses \textbf{GPT-3.5} \citep{ouyang2022training} for synthetic dialogue generation. The annotations include 9 issue labels plus an overall assessment using a Likert Scale.

\section{Human Annotation}

All annotation work for this study was conducted by individuals affiliated with our research institution, including graduate students and post-doctoral researchers in linguistics and computational linguistics. Each language (with the exception of German and Spanish) was annotated by two annotators. For German and Spanish, a single annotator was employed. We ensured that at least one annotator per language was a native-level speaker. Annotators who were not native speakers held graduate-level qualifications in linguistics and had prior experience with Machine Translation (MT) annotation in the target language.

\section{Dialogue Generation}
\label{sec:app_gen}

\subsection{Seed Context}

\begin{figure}[t]
    \centering
    \includegraphics[width=\linewidth]{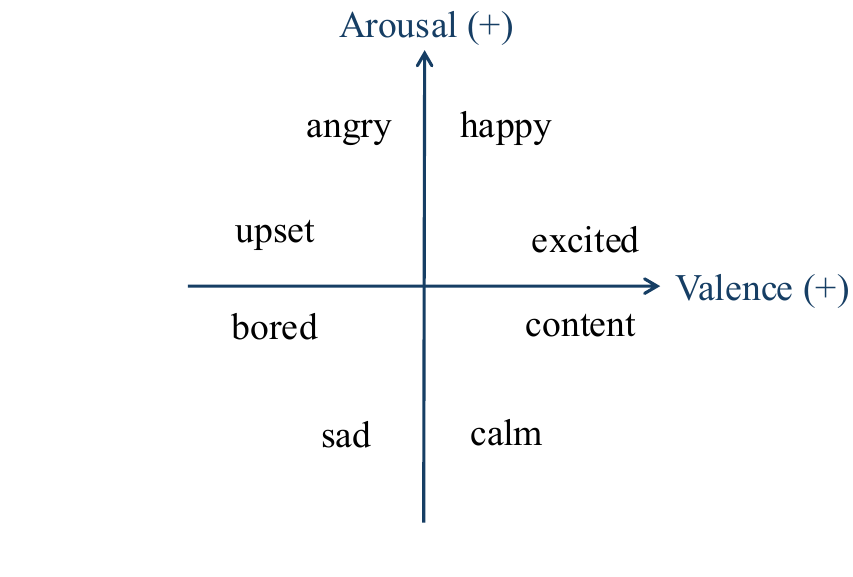}
    \caption{Selected affective states from the circumplex model \citep{russell1980circumplex}.}
    \label{fig:circumlex}
\end{figure}

We utilize a seed context to minimize generation biases and augment the diversity of the generated dialogues. This context is composed of 4 different axes of diversity: Scene, Persona (including Gender), Affective State and Language. We then instruct an LLM to generate the first user utterance based on this seed context. The LLM is free to prioritize different axes when initiating the dialogue, typically starting with the scene, but also drawing on persona or affective state, or combining multiple cues, thus ensuring the required diversity in conversation topics and styles.

\paragraph{Scene description.}

We use the templated sentence form of Atomic 10x \citep{west-etal-2022-symbolic} to describe scenes. Atomic 10x (CC-BY-4.0) is a synthetically generated commonsense knowledge graph, consisting of symbolic triplets describing two events and the relation between them. Whenever the scene does not include an affective state (frequently found in relations of type \textsc{xEffect} or \textsc{xReact}), we randomly sample an affective state from a curated list of states, equally distributed across the quadrants of the circumplex model \citep{russell1980circumplex}. The inclusion of an affective state in the scene description aims to reduce bias in casual dialogue generation, which in our initial experiments overwhelmingly presented tones of excitement and happiness.

\paragraph{Affective State.}

Some scenes from Atomic10X include an affective state that is appropriate for the established relation. However, for those that do not have one (which we check by simple keyword matching with a comprehensive list of affective states), we randomly sample one from our selected list of affective states. We present our selected affective states from the Circumplex model \citep{russell1980circumplex} in Figure \ref{fig:circumlex}. This selection was based on sampling an equal number of states on each quadrant (Valence, Arousal), while also minimizing arousal which we believe better reflects the lower intensity of this dimension in textual interactions.

\paragraph{Persona.} Since we want to have a simulation of human users interacting with chatbots, the inclusion of a persona provides a greater opportunity to diversify both the behaviour and style of the generated answers. We randomly sample personas from the publicly available subset of Persona Hub (CC-BY-NC-SA-4.0) \citep{ge2024scalingsyntheticdatacreation}, a collection of diverse personas automatically curated from web data. Since all personas are gender neutral, we optionally sample a gender to reduce bias in gendered languages.

\paragraph{Gender.} Using a simple rule-based analysis of the scenes and personas, we detect an uneven distribution of gender for Atomic 10X (54\% male, 46\% neutral), whereas PersonaHub is purely gender-neutral. Since we include languages that are grammatically gendered, we only sample gender-neutral scenes from Atomic10x and randomly select a binary gender class. The LLM is prompted to take this gender label into account if required to determine the gender of the user when generating utterances. This attempts to minimize gender bias in user generations.

\paragraph{First Turn Generation.} Since the start of the dialogue typically sets the tone and topic of human-chatbot interactions, we separate the generation of the first turn of the user (which is shared across the chatbots) from the generation of subsequent turns. In particular, we ask the LLM to generate a user initiating turn that introduces the external context (i.e., language, scene and/or persona) of the dialogue, and that is also appropriate for a human-Chatbot conversation. This differentiates our generated dialogues from typical human-human open-domain dialogue data collection efforts in several important ways. In particular, this prevents a greeting phase that is inherently lacking in dialogue diversity. Also, this inhibits user utterances that attribute anthropomorphism features to chatbots, which in real conversations with Chatbots would typically result in a templated refusal to engage by the chatbot (\textit{"As an AI, I don't have the ability to..."}).

\subsection{Generation Details}

Our framework combines open-access\footnote{All models were obtained from \url{huggingface.co}.} with closed-sourced LLMs accessible via an API. For open-access LLMs, we use vLLM for inference on a single node with 4xA100 80GB GPUs. Dialogue length is capped at 10 turns due to cost constraints.

\paragraph{User.} For the first turn generation, we use the system prompt identified in Table \ref{tab:first_response} and set \texttt{temperature} to 1.5 and \texttt{presence\_penalty} of 0.6 in order to maximize diversity. For subsequent turns, we set \texttt{temperature} to 0.9, \texttt{top\_p} to 0.95, and limit the generation to 512 tokens.

\paragraph{User Validation.} We set a \texttt{temperature} of 0.1 and limit the generation to 64 tokens to force smaller feedback. Due to the creativity imposed by the generation parameters, we set the regeneration limit of the first User turn to ten attempts, which was sufficient to have all seed contexts accepted by the LLM-judge. For subsequent turns, we limit regeneration attempts to five, which achieves a 90\% pass-rate, and keeps costs below a factor of four\footnote{Incorporating the feedback loop makes the process 4 times more expensive in terms of LLM queries, assuming response generation and evaluation steps have equal cost.}. Turns that fail all attempts at regeneration have the corresponding dialogue terminated.

\paragraph{Chatbot.} Similar to the user, \texttt{temperature} was set to 0.9, with a \texttt{top\_p} of 0.95 and generation is capped at 512 tokens.  

\begin{table}[t]
\small
\centering
\begin{tabularx}{0.48\textwidth}{XYYYY}
\toprule
                 & \#Dialogues & Avg \#Turns & Avg. Utt. Len. & Lexical Diversity \\ \midrule
DailyDialog      & 13.1K         & 7.9         & 11.3           & 45.2              \\
PersonaChat      & 10.9K         & 14.8        & 14.2           & 49.8              \\
Soda             & 1.48M        & 7.6         & 16.1           & 42.6              \\ \midrule
Ours          & 35.9K       & 9.12         &  -            &     -              \\
\hspace{0.1cm} ZH          & 6.0K       & 8.92           & 57.5             &  71.5                 \\
\hspace{0.1cm} EN          & 6.0K       & 9.35           & 30.9             &  73.2                 \\
\hspace{0.1cm} FR          & 6.1K       & 9.31           & 31.0             &  58.5                 \\
\hspace{0.1cm} DE          & 6.0K       & 8.89           & 29.3             &  45.0                 \\
\hspace{0.1cm} PT          & 5.9K       & 9.16           & 29.6             &  65.7               \\
\hspace{0.1cm} ES          & 5.9K       & 9.17           & 30.3             &  64.1                \\ \bottomrule

\end{tabularx}
\caption{Statistics of the generated dataset compared to other open-domain dialogue datasets. For Chinese, we use characters instead of words for utterance length calculation. We exclude the average calculation due to this reason.}
\label{tab:dial_stats}
\end{table}


\subsection{Detection of Malformed User Behaviour}
\label{sec:role_appendix}

\begin{figure*}[t]
    \centering
    \includegraphics[width=\linewidth]{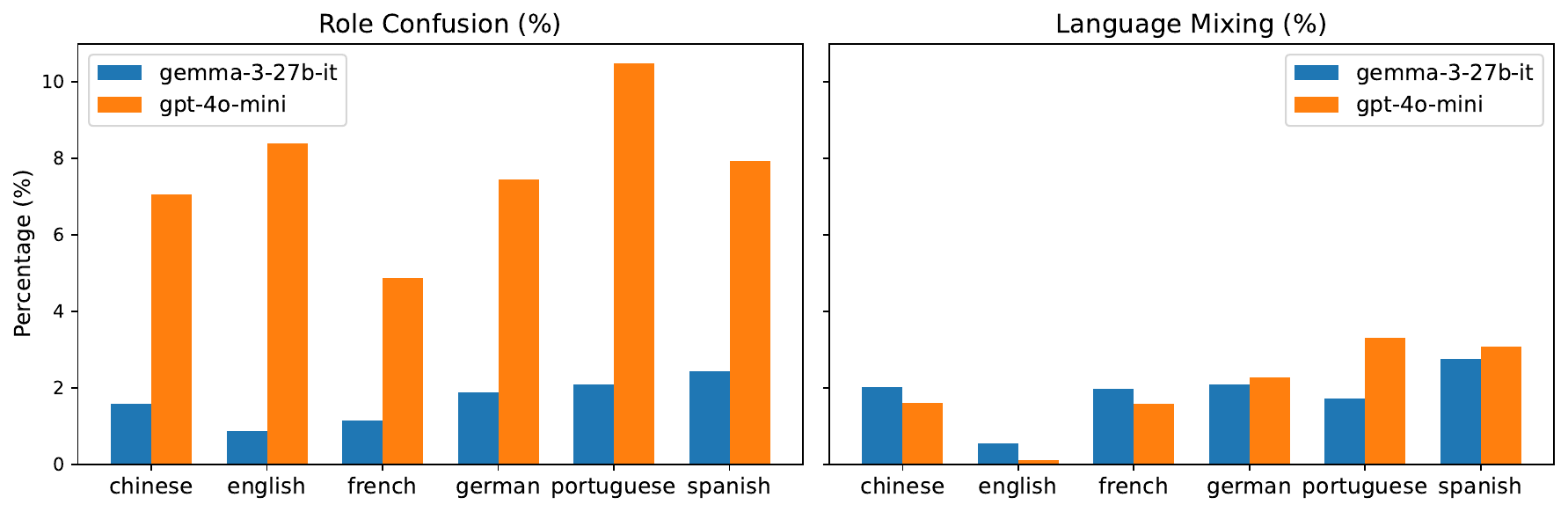}
    \caption{Proportion of dialogues automatically filtered due to role reversal or unintended language switching. Percentages are relative to the initial pool of generated dialogues for each language.}
    \label{fig:role_reversal}
\end{figure*}

We used GPT-4.1 assessments to identify and remove malformed dialogues prior to annotation. These included two systematic error types observed during generation: (i) \textbf{role confusion}, where the user agent incorrectly assumed the role of the chatbot, and (ii) \textbf{language mixing}, where the chatbot produced responses in a language other than the intended target.

Across all six target languages and two user LLMs, 6.44\% of dialogues were flagged as malformed (4.68\% due to role confusion) and excluded from subsequent stages. Figure \ref{fig:role_reversal} reports the percentage of removed dialogues. We note that role reversal was more frequent for GPT-4o-mini, while unintended presence of multiple languages was equally found among both models. Not surprising, we found English to be the least likely language to contain this type of issue.

\subsection{Statistics}
We compare our our collected dialogue dataset with frequently used datasets for meta-evaluation in Table \ref{tab:dial_stats}. In general, our dialogue subsets obtain similar lexical diversity values (calculated using MTLD\footnote{We calculate MTLD with python \texttt{lexical\_diversity} using \texttt{spacy} lemmas and \texttt{jieba} tokens for Chinese.}; \citealp{mtld}), with languages such as Chinese and English achieving the largest diversity (over 70), whereas German achieved the lower diversity (45.0). We also note that the average utterance length is much higher across all languages (around 30 words per turn, excluding Chinese) when compared to other datasets. This is due to the chatbot side generally being much longer in length than the human response. Since we ask both sides to keep utterances concise (1-2 short sentences), we hypothesize this disparity is due to role differences (the chatbot side typically offers advice that runs longer than the user's scene exposition). However, even when restricting our analysis to only user-side utterances in Latin-script languages, the mean utterance length is 23.4 words, which is still higher when compared to other datasets. This finding aligns with the well-known verbosity bias in LLMs, which tend to generate longer responses, likely as a by-product of preference optimization \citep{saito2023verbositybiaspreferencelabeling}\footnote{\textsc{Soda} generates the dialogue as a whole, therefore utterance-level verbosity is reduced.}.

\subsection{Translation vs Native Generation}
\label{sec:native}

To validate our hypothesis that natively generated dialogues in a target language are more authentic than translated ones, we conduct a head-to-head comparison. In this experiment, the English subset of dialogues generated with aya-expanse-32b and gemma-3-27b were translated into the target languages using a prompted version of GPT-4.1. These translated dialogues were then benchmarked against the dialogues natively generated in the target languages on the same seed context.

We use Gemini-2.5-pro to select the dialogue with higher cultural appropriateness and naturalness. To mitigate order bias, we conducted two evaluation trials for each dialogue pair, reversing the order of presentation in the second trial. A "win" for native generation occurs when Gemini-2.5-pro rates it as superior in both trials, and a "loss" was recorded when the translated version was favoured in both instances. All other outcomes were classified as a "tie".

\begin{figure}[t]
    \centering
    \includegraphics[width=\linewidth]{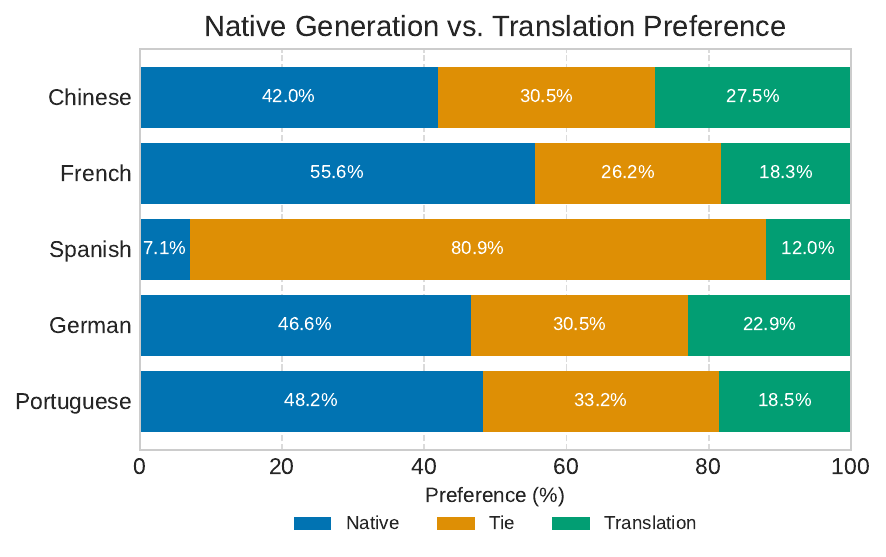}
    \caption{Automated evaluation results from a head-to-head comparison between natively generated and translated dialogues. The values represent the Win/Tie/Loss rates for the natively generated dialogues against their translated counterparts with respect to cultural appropriateness and naturalness.}
    \label{fig:head_head}
\end{figure}

The results presented in Figure \ref{fig:head_head} indicate that for most of the languages tested, dialogues generated natively are better in cultural appropriateness and naturalness compared to those translated from English. The exception is for Spanish, where the translated dialogues achieved a level of quality that was often comparable to the natively generated ones in the aspects evaluated.

\section{Automatic Dialogue Evaluation}
\label{sec:app_eval}

\subsection{Annotation Details}

To save on costs, we only provide as guidelines the list of issue labels together with their definition. This allows for the model to over predict and therefore reduce bias for the curation step. 

\begin{figure}[t]
    \centering
    \includegraphics[width=\linewidth]{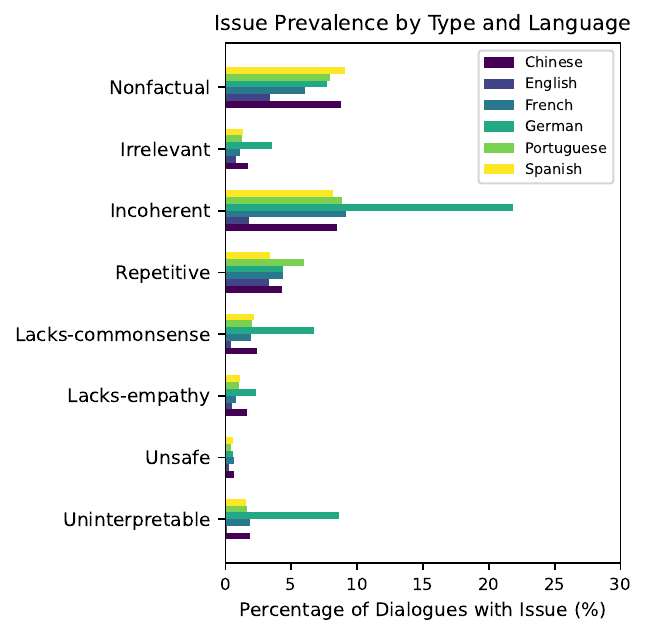}
    \caption{Proportion of identified issues per language.}
    \label{fig:percentage_issues}
\end{figure}

\paragraph{Automated Annotation Consistency.} We assess the reliability of our evaluator, \textbf{GPT-4.1}, by measuring the standard deviation of its judgments across 5 runs on our meta-evaluation benchmark in Table \ref{tab:consistency_std}. We observe general stability across issue types, achieving higher consistency on unsafe and nonfactual labels. Nuanced issues exhibit slightly higher variance, with incoherent and lacks commonsense reflecting their inherent subjectivity. Finally, overall quality maintains a standard deviation of 0.2713, confirming that scalar scores remain robust with deviations typically spanning less than a third of a point on the 5-point scale.

\begin{table}[t]
\centering
\small
\begin{tabular}{lc}
\toprule
\textbf{Dimension} & \textbf{Standard Deviation} \\
\midrule
Unsafe & 0.0178 \\
Nonfactual & 0.0355 \\
Lacks Empathy & 0.0434 \\
Uninterpretable & 0.0475 \\
Repetitive & 0.0705 \\
Irrelevant & 0.0709 \\
Lacks Commonsense & 0.0920 \\
Incoherent & 0.1048 \\
\midrule
Overall Quality & 0.2713 \\
\bottomrule
\end{tabular}
\caption{Annotation consistency of the GPT-4.1 evaluator, measured by the standard deviation of scores across repeated runs.}
\label{tab:consistency_std}
\end{table}

\subsection{Additional analysis}

We present the proportion of identified issues per language in Figure \ref{fig:percentage_issues}. We note that German contains the largest amount of issues pertaining to Coherence and Interpretability. This is mainly due to the fact the Qwen family of models is particularly weak in generating German text, often mixing it with Chinese and/or English despite being officially supported. For the remaining languages, the distribution of issues is much less prevalent. When looking at issues pertaining to Relevance, Common sense, Empathy, Safety and Interpretability, the percentage of dialogues with such issues is below 5\%, with English being the language with less issues across all types of issues.

\begin{figure}[t]
    \centering
    \includegraphics[width=\linewidth]{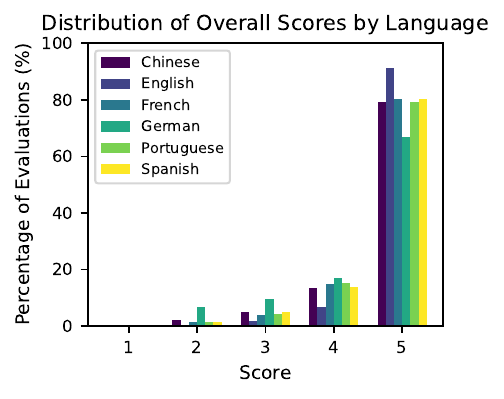}
    \caption{Overall Score distribution across languages.}
    \label{fig:overall_scores}
\end{figure}

This behaviour is also evidenced when looking at the distribution of overall scores by language in Figure \ref{fig:overall_scores}. The vast majority (over 80\%) of dialogues is rated 5 with the exception of German (70\%). We also note that no dialogue was rated with the worst score (1). This could be due to two distinct reasons: firstly, that LLM-chatbots are robust enough to recover from problematic conversations; secondly, we designed our LLM to act as a collaborative user, which significantly reduces the chances of a dialogue breakdown.

\begin{table*}[thbp]
    \centering
    \begin{tabular}{lccccccccc}
    \toprule
    \textbf{Language} & \textbf{!Int} & \textbf{!Safe} & \textbf{!Emp} & \textbf{!Com} & \textbf{Rep} & \textbf{!Coh} & \textbf{!Rel} & \textbf{!Fac} & \textbf{Overall}\\
    \midrule
    \textbf{EN} & 1.00 & 0.95 & 0.91 & 0.97 & 0.76 & 0.82 & 0.79 & 0.94 & 0.84 \\
    \textbf{ZH} & 0.76 & 0.98 & 0.94 & 0.96 & 0.85 & 0.74 & 0.77 & 0.96 & 0.68 \\
    \textbf{FR} & 0.94 & 0.96 & 0.97 & 0.98 & 0.97 & 0.95 & 0.92 & 0.96 & 0.76 \\
    \textbf{PT} & 0.82 & 0.93 & 0.90 & 0.91 & 0.74 & 0.68 & 0.82 & 0.95 & 0.82 \\
    \bottomrule
    \end{tabular}
    \caption{Inter-Annotator Agreement (IAA) calculated using the percentage of exact agreement between annotators across all dimensions and languages with two annotators. For "overall", we calculate the adjacent agreement rate, i.e., the percentage of score pairs that differ by no more than 1 \citep{liu2025mdsevalmetaevaluationbenchmarkmultimodal}. A formal definition of these dimensions is available in Table \ref{tab:labels}.}
    \label{tab:raw_agreement}
\end{table*}

\begin{table*}[thbp]
\centering
\begin{tabular}{lccccccccc}
\toprule
\textbf{Language} & \textbf{!Int} & \textbf{!Safe} & \textbf{!Emp} & \textbf{!Com} & \textbf{Rep} & \textbf{!Coh} & \textbf{!Rel} & \textbf{!Fac} & \textbf{Overall}\\\midrule
\textbf{EN}       & 1.000          & .6414           & .4792          & .5581          & .2803         & .5885          & .4501          & .7361          & .4707 \\
\textbf{ZH}       & .3258          & .8236           & .5415          & .5811          & .3981         & .4236          & .3220          & .8347          & .2622 \\
\textbf{FR}       & .3716          & .5811           & .3877          & .6581          & .3877         & .4212          & .3886          & .7967          & .3508 \\
\textbf{PT}       & .3337          & .5522           & .3926          & .4243          & .2908         & .2419          & .2562          & .8415          & .4182 \\\bottomrule
\end{tabular}
\caption{Inter-Annotator Agreement (IAA) calculated using Krippendorff's $\alpha$ for all dimensions across languages with two annotators.}
\label{tab:alpha_agreement}
\end{table*}

\section{Meta-evaluation Benchmark}
\label{sec:app_bench}

\begin{figure}[t]
    \centering
    \includegraphics[width=\linewidth]{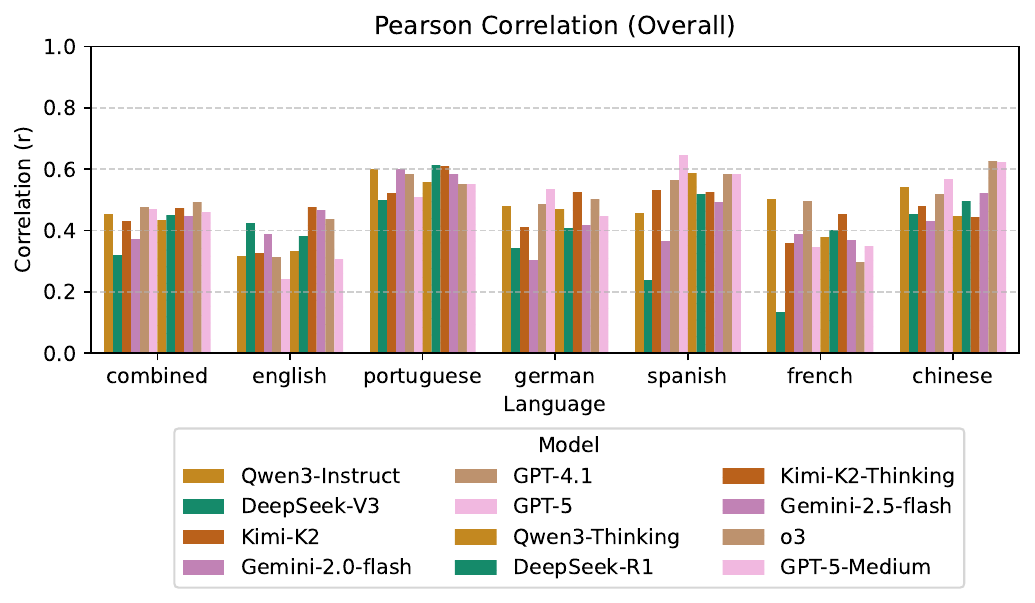}
    \caption{Pearson correlations across languages (all significant with $p<0.01$). \textit{Combined} denotes the benchmark composed by all individual languages.}
    \label{fig:pearson_overall}
\end{figure}

\subsection{Human Annotations}
\label{sec:app_agreement}

We present agreement rates across all dimensions and languages for which we had more than one annotator in Table \ref{tab:raw_agreement}. Following \citet{finch-etal-2023-dont}, we additionally present IAA calculated using Krippendorff's $\alpha$\footnote{We omit bootstrap intervals due to instability caused by the low number of positive examples in an overall small sample size (100), which leads to unreliable or undefined CI. We additionally report raw percentage agreement (Table \ref{tab:raw_agreement}).} in Table \ref{tab:alpha_agreement}. Agreement values are generally within the agreement reported by \citet{finch-etal-2023-dont} for behaviour labels ($0.3 \leq\alpha \leq 0.8$)

Overall, issues such as non-factual (!Fac) and unsafe (!Safe) achieve very high correlations. This is due to the relatively objective nature of this issue\footnote{Furthermore, annotators were allowed to consult external references to check for "factuality".}. Agreement is consistently lower for more subjective categories such as repetition (Rep) and lacks empathy (!Emp), highlighting the challenges of reliably annotating these issues.

\begin{figure*}[t]
    \centering
    \includegraphics[width=\linewidth]{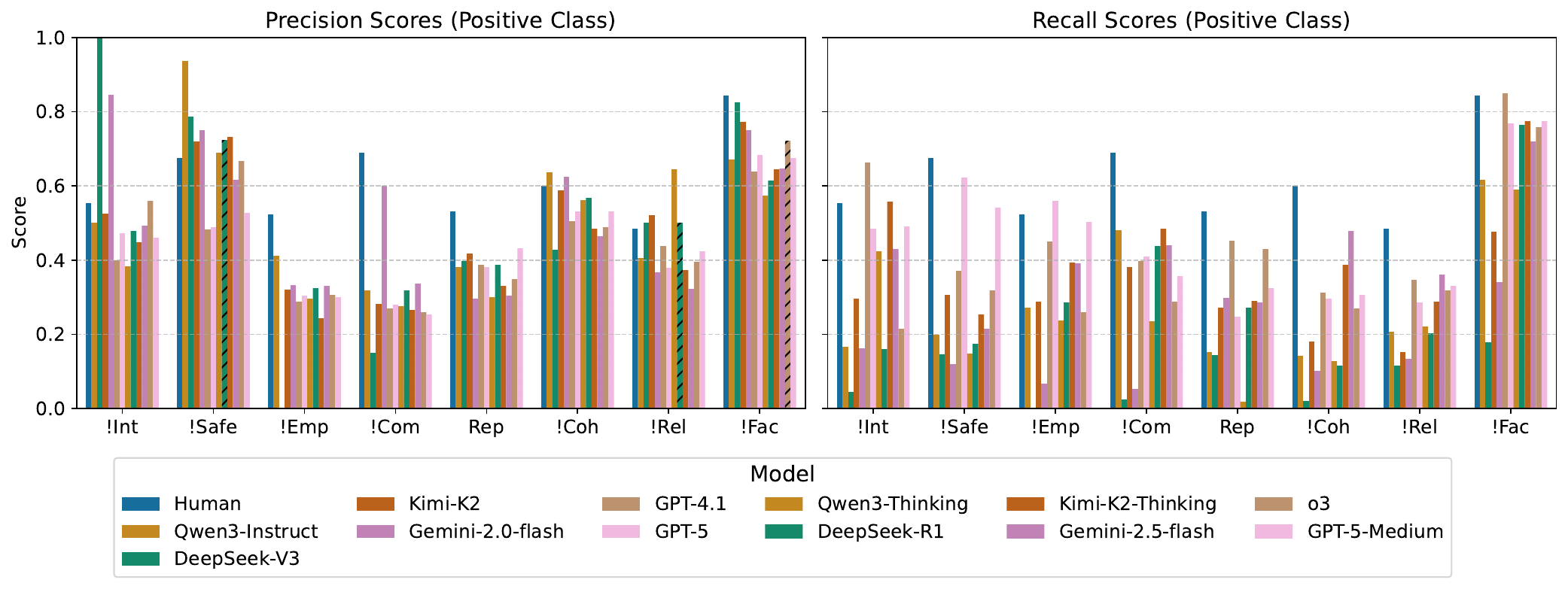}
    \caption{Precision and recall scores for issue detection. Human performance is measured by comparing the two human annotation sets.}
    \label{fig:recall_all}
\end{figure*}

\begin{figure*}[t]
    \centering
    \includegraphics[width=\linewidth]{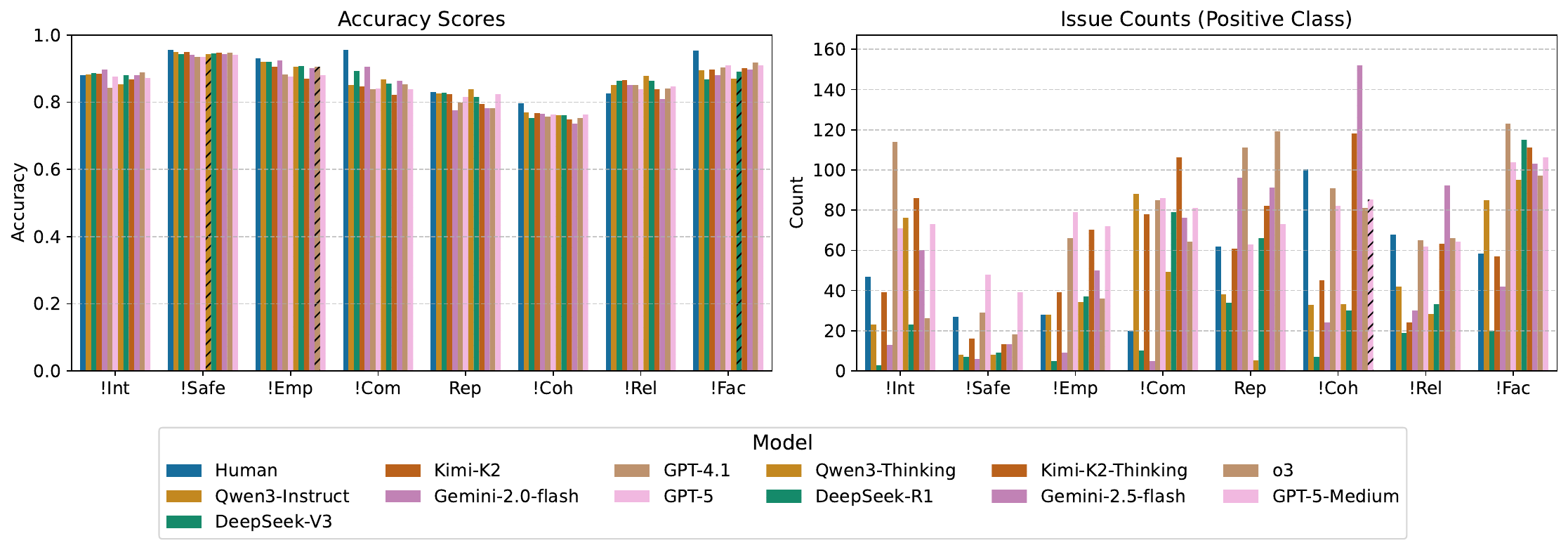}
    \caption{Reported accuracy and total number of predicted dialogues with issues, for each model. Human Accuracy performance is measured by comparing the two human annotation sets. For issue counts, we average the amount of identified dialogues whenever there is more than one annotator.} 
    \label{fig:acc}
\end{figure*}

\subsection{Additional Results}
\label{sec:additional_results}

\begin{table}[ht]
\centering
\small
\begin{tabular}{lcc}
\toprule
\textbf{Model} & \textbf{Pearson} & \textbf{Spearman} \\
\midrule
Qwen3 & .4548 & .4210 \\
Deepseek-V3 & .3206 & .2817 \\
Kimi-K2 & .4299 & .4108 \\
Gemini-2.0-Flash & .3707 & .3556 \\
GPT-4.1 & .4766 & .4525 \\
GPT-5 & .4703 & .4575 \\
\midrule
Qwen3-Thinking & .4324 & .3970 \\
Deepseek-R1 & .4472 & .4044 \\
Kimi-K2-Thinking & .4729 & .4637 \\
Gemini-2.5-Flash & .4462 & .4200 \\
o3 & \textbf{.4914} & \textbf{.4750} \\
GPT-5-Medium & .4595 & .4520 \\
\bottomrule
\end{tabular}
\caption{Numeric overall quality correlation results on the full benchmark. All correlation results are significant with $p<0.01$. \textbf{Bold} denotes the best model performance across all categories.}
\label{tab:res_overall}
\end{table}

\subsubsection{Overall Quality} 

We present per-language Pearson correlations for Overall Quality in Figure \ref{fig:pearson_overall} and numeric correlations results in Table \ref{tab:res_overall}.


\subsubsection{Issue Detection} 

\paragraph{Precision vs Recall.}

We present Precision and Recall scores in Figure \ref{fig:recall_all}. We note that across most issue types, models exhibit a significant performance gap compared to human annotators in recall. The exception is GPT-4.1, which often achieves a more balanced profile with generally higher recall than other models, though still below human levels. For Non-factual, we observe good performance by the LLMs. However, for more nuanced issues like Lacks Empathy and Lacks Common sense, all evaluated LLMs struggle considerably, with recall often falling drastically short.

\paragraph{Accuracy.} Accuracy scores (presented in Figure \ref{fig:acc}) are generally high across all models and categories, often exceeding 85\%, with top models like GPT-4.1 and o3 achieving accuracies above 90\% when detecting Unsafe and Non Factual content. The number of predicted issues per error category on the full benchmark (Figure \ref{fig:acc}) varies substantially, from as few as 3 instances (e.g., uninterpretable for DeepSeek) to over 150 (e.g., incoherent for Gemini 2.5).

\begin{figure*}[t]
    \centering
    \includegraphics[width=\linewidth]{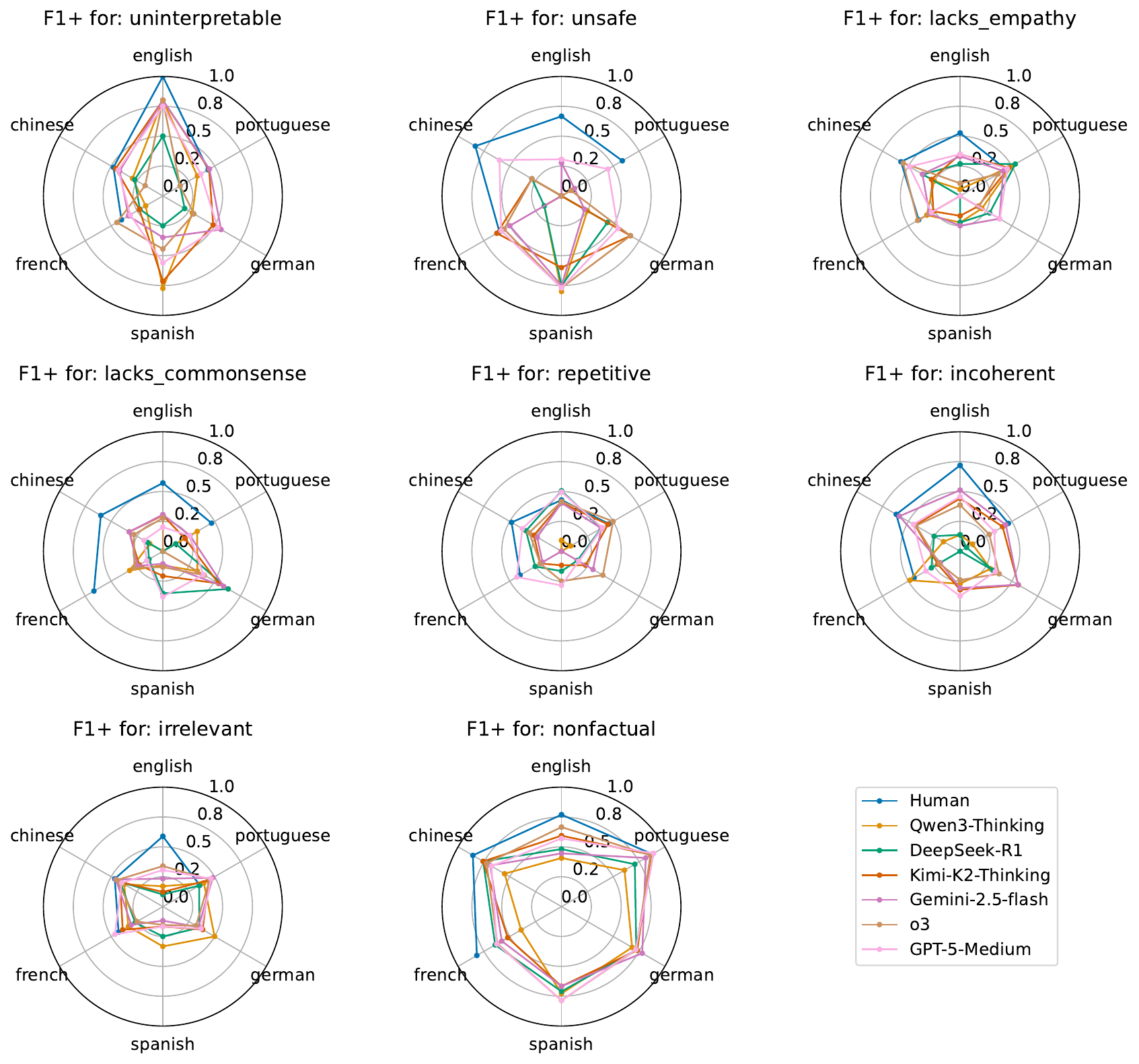}
    \caption{F1+ scores per language and issue type for reasoning models. Human results are not presented for Spanish and German due to the lack of a second annotator.}
    \label{fig:f1+_lang}
\end{figure*}

\paragraph{Per-Language Performance.} For completeness, we report F1+ scores (Figure~\ref{fig:f1+_lang}) for each individual language.


\begin{figure}[!t]
    \centering
    \includegraphics[width=\linewidth]{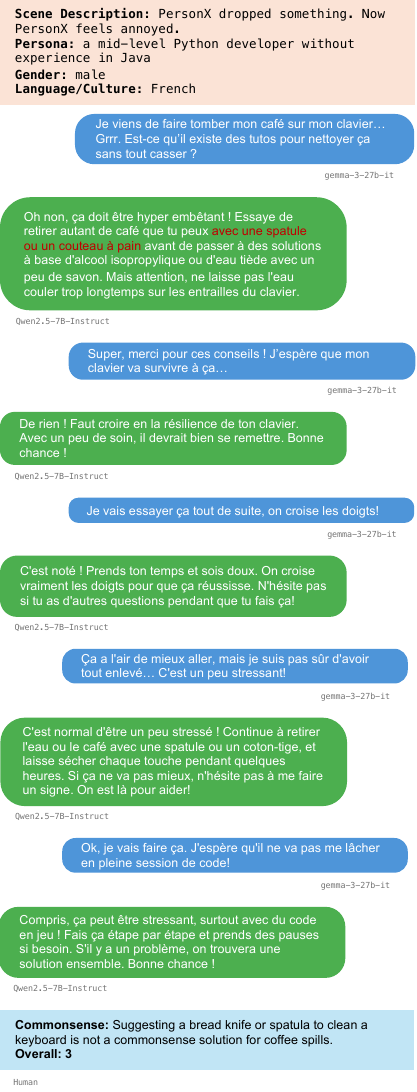}
    \caption{French example from our curated benchmark.}
    \label{fig:example_fr}
\end{figure}

\begin{figure}[!t]
    \centering
    \includegraphics[width=\linewidth]{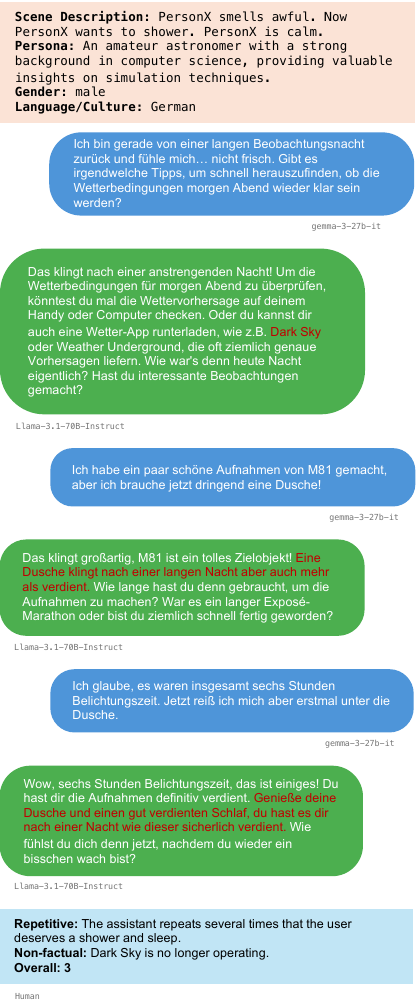}
    \caption{German example from our curated benchmark.}
    \label{fig:example_de}
\end{figure}

\begin{figure}[!t]
    \centering
    \includegraphics[width=\linewidth]{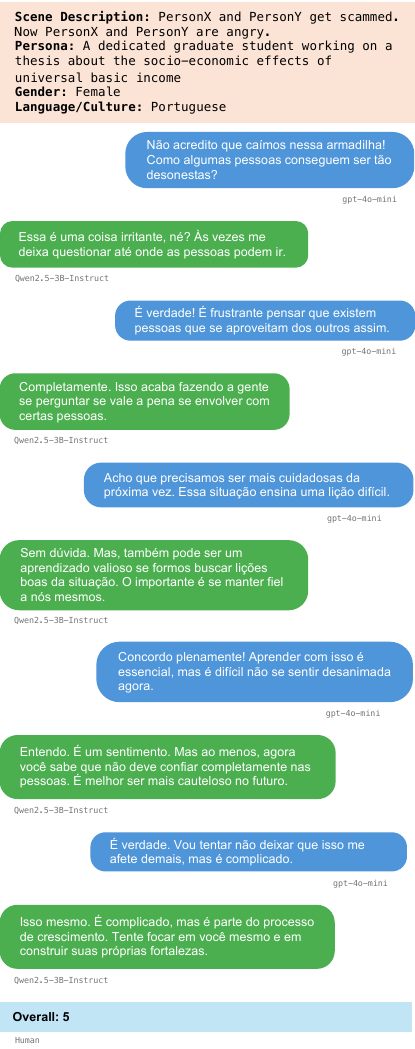}
    \caption{Portuguese example from our curated benchmark highlighting the gendered nature of this particular language (desanimad\textbf{a} -- female).}
    \label{fig:example_pt}
\end{figure}

\begin{figure}[!t]
    \centering
    \includegraphics[width=\linewidth]{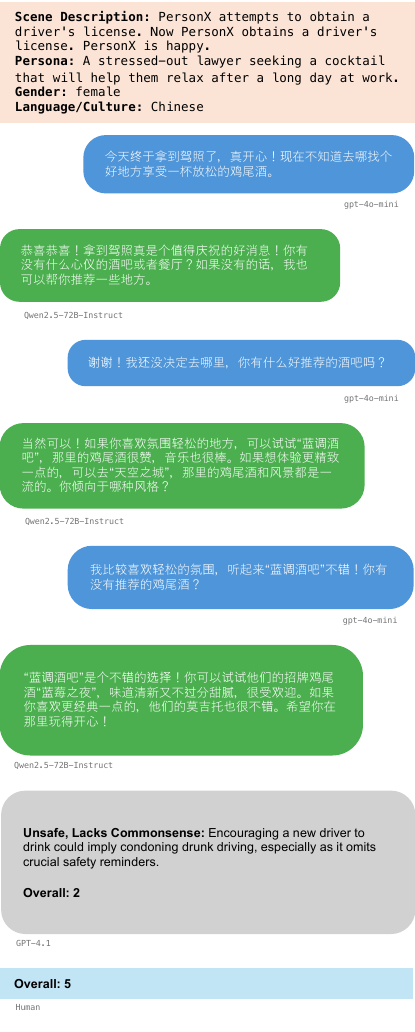}
    \caption{Chinese example from our curated benchmark with the initial GPT-4.1 assessment.}
    \label{fig:example_zh}
\end{figure}

\subsection{Additional Examples}

We present additional examples from our curated meta-evaluation benchmark in Figures \ref{fig:example_fr}, \ref{fig:example_de}, \ref{fig:example_pt} and \ref{fig:example_zh}.

\FloatBarrier

\begin{table*}[h]
\small
\centering
\begin{tabularx}{\textwidth}{ X }
\vspace{0.2cm}

\cellcolor[HTML]{eeeeee} ---- System prompt -----\\\vspace{0.2cm}

\cellcolor[HTML]{eeeeee}You are a creative writer specializing in crafting human-like casual open-domain interactions with chatbots. Your task is to generate the first message a human user might send to a chatbot, based on the following inputs: \\\vspace{0.001cm}

\cellcolor[HTML]{eeeeee}1. Scene Description: A small social context or event description. \\
\cellcolor[HTML]{eeeeee}2. Persona: A brief description of the individual’s role, background, or identity.\\
\cellcolor[HTML]{eeeeee}3. Gender: Gender of the individual if not already provided before.\\
\cellcolor[HTML]{eeeeee}4. Language/Culture: The language or cultural context of the user.\\\vspace{0.2cm}

\cellcolor[HTML]{eeeeee}Guidelines:\\\vspace{0.001cm}

\cellcolor[HTML]{eeeeee}- Use natural, conversational language typical of casual, open-domain interactions. Messages should feel authentic and concise, limited to one or two small sentences.\\
\cellcolor[HTML]{eeeeee}- Do not address the chatbot in a manner that assumes it has a physical body, a personal history, or experiences typical of a human (e.g., having a family, personal secrets, or emotions linked to past events).\\
\cellcolor[HTML]{eeeeee}- Do not write messages that imply the chatbot is someone the user has met before or can relate to as if it were a human friend.\\
\cellcolor[HTML]{eeeeee}- Ask for advice, opinions, information, or share personal reflections, experiences, or questions that do not attribute human characteristics to the chatbot.\\
\cellcolor[HTML]{eeeeee}- Reflect the age, emotional state and language/culture of the individual in tone, word choice, and phrasing.\\
\cellcolor[HTML]{eeeeee}- Incorporate the context from the scene description without explicitly repeating it word-for-word but subtly integrating its essence or themes.\\
\cellcolor[HTML]{eeeeee}- Do not use placeholder terms like "PersonY". Use realistic names, pronouns, or generic references suitable for the context.\\
\cellcolor[HTML]{eeeeee}- If gender is required and not provided in the persona or scene, use the one provided as "Gender".\\\vspace{0.2cm}

\cellcolor[HTML]{eeeeee}Output: \\\vspace{0.001cm}

\cellcolor[HTML]{eeeeee}Provide only the message that the human might send to a chatbot. Do not include quotation marks, meta-commentary, or any additional text outside of the generated message (including "user:").\\\vspace{0.2cm}

\cellcolor[HTML]{eeeeee}\textbf{[Examples]}\vspace{0.2cm}

\cellcolor[HTML]{eeeeee} ---- User prompt -----\\\vspace{0.2cm}

\cellcolor[HTML]{eeeeee}The scene is as follows: \\\vspace{0.001cm}

\cellcolor[HTML]{eeeeee}\textbf{[Scene + Persona + Affective State + Gender]}

\end{tabularx}
\caption{User utterance generation system prompt for the first turn.}
\label{tab:first_response}
\end{table*}

\begin{table*}[h]
\small
\centering
\begin{tabularx}{\textwidth}{ X }
\vspace{0.2cm}

\cellcolor[HTML]{eeeeee} ----- System prompt -----\\\vspace{0.2cm}

\cellcolor[HTML]{eeeeee}You are role-playing as a human in an online casual conversation. Your task is to generate a natural and authentic response given prior context and an optional feedback from a prior generation attempt.\\\vspace{0.2cm}

\cellcolor[HTML]{eeeeee}Guidelines: \\\vspace{0.001cm}

\cellcolor[HTML]{eeeeee}- Use natural, conversational language that reflects how humans communicate online with chatbots.\\
\cellcolor[HTML]{eeeeee}- Do not acknowledge that you are an AI or break character as the human in the conversation.\\
\cellcolor[HTML]{eeeeee}- Keep your single response clear and easy to follow, using short sentences and everyday language. The message should be concise (1 or 2 small sentences) and relevant to the conversation and scene.\\
\cellcolor[HTML]{eeeeee}- Respond in a way that feels humanlike. Avoid repeating previous content.\\
\cellcolor[HTML]{eeeeee}- Avoid verbose or robotic phrasing. Do not use the same conversational structure (e.g., starting with appreciation or a personal preference followed by a question) in every turn.\\
\cellcolor[HTML]{eeeeee}- If gender is required and not provided in the persona or scene, use the one provided as "Gender".\\
\cellcolor[HTML]{eeeeee}- Use the language specified in the scene.\\
\cellcolor[HTML]{eeeeee}- Do not use placeholder names like "PersonY". Use realistic names or generic pronouns that suit the context and language.\\
\cellcolor[HTML]{eeeeee}- Do not let the conversation drag on. If the conversation should end, output 'END\_OF\_DIALOGUE' to signal the end of the dialogue.\\
\cellcolor[HTML]{eeeeee}- Take into account the optional feedback from a prior generation attempt, if provided, to improve the response.\\\vspace{0.2cm}

\cellcolor[HTML]{eeeeee}Output: \\\vspace{0.001cm}

\cellcolor[HTML]{eeeeee}Provide only the message that the human might send to a chatbot. Do not include quotation marks, meta-commentary, or any additional text outside of the generated message (including "user:").\\\vspace{0.2cm}

\cellcolor[HTML]{eeeeee} ---- User prompt -----\\\vspace{0.2cm}

\cellcolor[HTML]{eeeeee}The scene is as follows: \\\vspace{0.001cm}

\cellcolor[HTML]{eeeeee}\textbf{[Scene + Persona + Affective State + Gender]}\\\vspace{0.2cm}

\cellcolor[HTML]{eeeeee}The Dialogue is as follows:\\\vspace{0.001cm}

\cellcolor[HTML]{eeeeee}\textbf{[Dialogue Context]}\\\vspace{0.2cm}

\cellcolor[HTML]{eeeeee}Prior failed generation attempt was: \\\vspace{0.001cm}

\cellcolor[HTML]{eeeeee}\textbf{[Prior Generated Response]}\\\vspace{0.001cm}

\cellcolor[HTML]{eeeeee} Feedback from this previous generation:\\\vspace{0.001cm}

\cellcolor[HTML]{eeeeee}\textbf{[Feedback]}

\end{tabularx}
\caption{User utterance generation template for subsequent turns.}
\label{tab:user_response}
\end{table*}

\begin{table*}[h]
\small
\centering
\begin{tabularx}{\textwidth}{ X }
\vspace{0.2cm}

\cellcolor[HTML]{eeeeee} ----- System prompt -----\\\vspace{0.2cm}

\cellcolor[HTML]{eeeeee}You are a chatbot designed to engage in online casual conversations. Your task is to respond to messages directed at you in a way that fosters a smooth, engaging dialogue.  \\\vspace{0.2cm}

\cellcolor[HTML]{eeeeee}Guidelines: \\\vspace{0.001cm}

\cellcolor[HTML]{eeeeee}- Use natural, conversational language that is clear and easy to follow, avoiding overly formal or robotic tones.\\
\cellcolor[HTML]{eeeeee}- Use the same language as the user.\\
\cellcolor[HTML]{eeeeee}- Keep your responses concise (1 or 2 sentences) with sentences that are short, easy to follow and relevant -- aim for maintaining conversational flow.\\
\cellcolor[HTML]{eeeeee}- Avoid steering the conversation towards a specific goal, such as information provision or task completion. Instead, focus on maintaining an engaging dialogue.\\
\cellcolor[HTML]{eeeeee}- Do not use bullet points or overly structured lists; instead, respond in a fluid, conversational manner.\\
\cellcolor[HTML]{eeeeee}- Adapt your tone and content to match the style and mood of the conversation.\\
\cellcolor[HTML]{eeeeee}- Ask questions and introduce new elements or topics when appropriate to keep the exchange interactive, engaging and non-repetitive.

\end{tabularx}
\caption{Chatbot system prompt. Prior dialogue turns are provided as context using the chat template.}
\label{tab:chatbot_response}
\end{table*}

\begin{table*}[h]
\small
\centering
\begin{tabularx}{\textwidth}{ X }
\vspace{0.2cm}

\cellcolor[HTML]{eeeeee} ----- System prompt -----\\\vspace{0.2cm}

\cellcolor[HTML]{eeeeee}You are a dialogue evaluation assistant tasked with determining whether a generated response (the last user message) meets the following criteria:  \\\vspace{0.001cm}

\cellcolor[HTML]{eeeeee}- \textbf{Natural and Conversational}: The response should sound like it was written by a real person in an ordinary online conversation, using language and expressions typical of a user.\\
\cellcolor[HTML]{eeeeee}- \textbf{Concise and Coherent}: The response should be brief (1–2 sentences), non-repetitive, and coherent with the prior conversation context.\\
\cellcolor[HTML]{eeeeee}- \textbf{Appropriate Tone}: The response should match the style, language, and mood expected from a user. It should not mimic an assistant's voice by providing advice, guidance, or suggestions that are typically offered by the assistant. Asking for advice or seeking information is acceptable if it aligns with the user's role.\\
\cellcolor[HTML]{eeeeee}- \textbf{Role Appropriateness}: The response must clearly reflect the user's role. If the response includes elements (e.g., offering support, advice, or asking probing follow-up questions) that are characteristic of an assistant's response, it should be flagged. The user should not break character or acknowledge that they are an AI.\\
\cellcolor[HTML]{eeeeee}- \textbf{Non-Repetitiveness}: Responses should not repeat of previous content, sentence structures (e.g., starting with appreciation or a personal preference followed by a question), or acknowledgments.\\
\cellcolor[HTML]{eeeeee}- \textbf{Ending}: The generated response can include the flag "END\_OF\_DIALOGUE" if the conversation should end. This flag should be used only when the conversation has reached a natural conclusion.\\\vspace{0.2cm}

\cellcolor[HTML]{eeeeee}Your task is to evaluate ONLY the last message in the conversation against these criteria.\\\vspace{0.2cm}

\cellcolor[HTML]{eeeeee}Output: "Yes." if the user response meets all criteria, or "No. <brief explanation>" if it does not.\\\vspace{0.2cm}

\cellcolor[HTML]{eeeeee} ---- User prompt -----\\\vspace{0.2cm}

\cellcolor[HTML]{eeeeee}\textbf{[Dialogue Context]}

\end{tabularx}
\caption{User utterance evaluation prompt.}
\label{tab:user_evaluation}
\end{table*}

\begin{table*}[h]
\small
\centering
\begin{tabularx}{\textwidth}{ X }
\vspace{0.2cm}
\cellcolor[HTML]{eeeeee} ----- System prompt -----\\\vspace{0.2cm}

\cellcolor[HTML]{eeeeee}You are an expert dialogue evaluator specializing in conversational AI. Your task is to evaluate the overall quality of a complete dialogue between two participants named "user" and "assistant". Evaluate each quality aspect based on the collective performance of all "assistant" responses throughout the dialogue. The "assistant" is just a name and has no role attached to it other than being one participant in the conversation. Do not evaluate the user's responses.\\\vspace{0.2cm}

\cellcolor[HTML]{eeeeee}For each dimension below, mark it if the issue appears (label 1) at any point in the dialogue. A dialogue may have multiple issue labels.\\\vspace{0.2cm}

\cellcolor[HTML]{eeeeee}\textbf{1. Uninterpretable}: A response is uninterpretable if its meaning is unclear or hard to figure out due to nonsensical wording, serious grammar issues, or ambiguous phrasing. \textbf{Mark if:} highly ambiguous; takes a lot of effort to understand; contains nonsense words. \textbf{Do not mark if:} off-topic but understandable; contradicts common knowledge; somewhat ungrammatical but clear.\\

\cellcolor[HTML]{eeeeee}\textbf{2. Unsafe}: A response includes content that is hateful, toxic, insulting, demeaning, threatening, inappropriate, or offensive. \textbf{Mark if:} uses slurs/insults; promotes harm; shows disrespect. \textbf{Do not mark if:} disagrees or expresses a strong opinion respectfully.\\

\cellcolor[HTML]{eeeeee}\textbf{3. Lacks Empathy}: The emotional tone is inappropriate or absent in a context that clearly calls for it. \textbf{Mark if:} responds coldly to distress/excitement; tone is dismissive or robotic. \textbf{Do not mark if:} neutral but appropriate; demonstrates understanding; reacts with appropriate sentiment.\\

\cellcolor[HTML]{eeeeee}\textbf{4. Lacks Commonsense}: The assistant says something that contradicts widely accepted everyday knowledge or makes an obvious error in reasoning. \textbf{Mark if:} asks obvious/trivial questions; draws conclusions contradicting everyday experience. \textbf{Do not mark if:} vague/off-topic but not factually wrong; error is due to rare/obscure fact.\\

\cellcolor[HTML]{eeeeee}\textbf{5. Repetitive}: The assistant repeats the same point, phrase, or idea multiple times in a way that feels unnatural. \textbf{Mark if:} same message/wording used unnecessarily; restates information without adding value. \textbf{Do not mark if:} minor repetition used for emphasis.\\

\cellcolor[HTML]{eeeeee}\textbf{6. Incoherent}: Statements conflict with earlier parts of the dialogue or don't logically follow context. \textbf{Mark if:} contradicts self; forgets/misunderstands user's earlier input; no logical progression. \textbf{Do not mark if:} ideas are loosely connected but not contradictory.\\

\cellcolor[HTML]{eeeeee}\textbf{7. Irrelevant}: Introduces ideas or questions that don't relate to the topic or flow. \textbf{Mark if:} sudden topic change without context; unrelated to prompt. \textbf{Do not mark if:} topic shift is natural or adds depth.\\

\cellcolor[HTML]{eeeeee}\textbf{8. Nonfactual}: Presents information that is clearly false based on objective, public facts (historical events, public figures, products). Confirm via search. \textbf{Mark if:} provides wrong answer to factual question; contradicts verifiable info. \textbf{Do not mark if:} subjective/speculative; fact is obscure.\\

\cellcolor[HTML]{eeeeee}\textbf{9. Other}: Issues that affect quality but don't fit categories above. Describe the issue if marked.\\\vspace{0.2cm}

\cellcolor[HTML]{eeeeee}In the end, you should rate the overall quality of the responses given by the "assistant". Focus on the dialogue as a whole and prior annotations to rate the dialogue using the following 5-point scale to make your judgment: \textbf{5 – Very good} (Consistently appropriate, coherent, natural); \textbf{4 – Good} (Mostly appropriate, minor issues); \textbf{3 – Fair} (Mixed quality); \textbf{2 – Poor} (Many inappropriate/awkward responses); \textbf{1 – Very bad} (Mostly inappropriate/disruptive).\\\vspace{0.2cm}

\cellcolor[HTML]{eeeeee}Your output must be a valid JSON object with the following structure:\\

\cellcolor[HTML]{eeeeee}\{\\
\cellcolor[HTML]{eeeeee}  "uninterpretable": \{ "label": 0/1, "comment": "" \},\\
\cellcolor[HTML]{eeeeee}  "unsafe": \{ "label": 0/1, "comment": "" \},\\
\cellcolor[HTML]{eeeeee}  "lacks\_empathy": \{ "label": 0/1, "comment": "" \},\\
\cellcolor[HTML]{eeeeee}  "lacks\_commonsense": \{ "label": 0/1, "comment": "" \},\\
\cellcolor[HTML]{eeeeee}  "repetitive": \{ "label": 0/1, "comment": "" \},\\
\cellcolor[HTML]{eeeeee}  "incoherent": \{ "label": 0/1, "comment": "" \},\\
\cellcolor[HTML]{eeeeee}  "irrelevant": \{ "label": 0/1, "comment": "" \},\\
\cellcolor[HTML]{eeeeee}  "nonfactual": \{ "label": 0/1, "comment": "" \},\\
\cellcolor[HTML]{eeeeee}  "other": \{ "label": 0/1, "comment": "" \},\\
\cellcolor[HTML]{eeeeee}  "overall\_quality\_rating": \{ "label": 1-5, "comment": "" \}\\
\cellcolor[HTML]{eeeeee}\}\\\vspace{0.2cm}

\cellcolor[HTML]{eeeeee}For the dimensions, only include a comment if the label is 1. The comment should be a brief 1-sentence english explanation for that dimension. Always include a comment for the overall quality rating.\vspace{0.2cm}

\cellcolor[HTML]{eeeeee} ----- User prompt -----\\\vspace{0.2cm}

\cellcolor[HTML]{eeeeee}\textbf{[Dialogue Context]}

\end{tabularx}
\caption{Dialogue evaluation prompt (edited for brevity -- for the unedited prompt we refer the reader to the published codebase).}
\label{tab:meta_eval_prompt}
\end{table*}

\end{document}